\documentclass{article}

\PassOptionsToPackage{numbers,sort&compress}{natbib}

\usepackage[eandd,final]{neurips_2026}
\usepackage[utf8]{inputenc}
\usepackage[T1]{fontenc}
\usepackage{courier}
\usepackage{hyperref}
\usepackage{url}
\usepackage{booktabs}
\usepackage{longtable}
\usepackage{needspace}
\usepackage{array}
\usepackage{etoolbox}
\usepackage{multirow}
\usepackage{amsfonts}
\usepackage{amsmath}
\usepackage{amssymb}
\usepackage{graphicx}
\usepackage{microtype}
\usepackage[table]{xcolor}
\definecolor{TableHeader}{HTML}{EAF0F5}
\definecolor{TableGroup}{HTML}{F2F6FA}
\definecolor{TableAccent}{HTML}{33539E}
\definecolor{TablePurple}{HTML}{A5678E}
\definecolor{TableFocusBlue}{HTML}{EDF2FA}
\definecolor{TableFocusPurple}{HTML}{F4EAF1}
\newcommand{\scorebest}[1]{\textcolor{TableAccent}{\textbf{#1}}}
\newcommand{\transferfocus}[1]{\textcolor{TablePurple}{\textbf{#1}}}

\usepackage{placeins}
\usepackage{tikz}
\usetikzlibrary{arrows.meta,positioning,fit,calc}
\usepackage[most]{tcolorbox}
\usepackage{listings}

\lstdefinelanguage{json}{
  basicstyle=\ttfamily\scriptsize,
  breaklines=true,
  showstringspaces=false,
  columns=fullflexible
}

\newtcolorbox{promptbox}[2][]{
  enhanced,
  breakable,
  colback=gray!4,
  colframe=gray!55,
  title={#2},
  fonttitle=\bfseries\small,
  boxrule=0.5pt,
  arc=2pt,
  left=4pt,
  right=4pt,
  top=4pt,
  bottom=4pt,
  #1
}

\newtcolorbox{codebox}[2][]{
  enhanced,
  breakable,
  colback=gray!3,
  colframe=gray!50,
  title={#2},
  fonttitle=\bfseries\small,
  boxrule=0.5pt,
  arc=2pt,
  left=4pt,
  right=4pt,
  top=4pt,
  bottom=4pt,
  #1
}
\title{\textsc{MedEvoEval}: Evaluating Continual Evolution of Doctor Agents through Simulated Clinical Episodes}

\author{%
  Hui Zhang\\
  Beijing Institute of Technology, Beijing, China\\
  \href{mailto:huizhang.ai.2025@gmail.com}{\texttt{huizhang.ai.2025@gmail.com}}
}

\hypersetup{
  hidelinks,
  pdftitle={MedEvoEval: Evaluating Continual Evolution of Doctor Agents through Simulated Clinical Episodes},
  pdfauthor={Hui Zhang}
}

\newcommand{\CodeURL}{https://github.com/huihuihenqiang/MedEvoEval}
\newcommand{\DataURL}{https://www.kaggle.com/datasets/anonymousmedevoeval/medevoeval-anonymous-review-dataset}

\newcommand{\CompetingInterestsStatement}{The author declares no competing interests.}
\newcommand{\ArtifactAvailability}{Code and data are available at \href{\CodeURL}{the code repository} and \href{\DataURL}{the dataset page}.}

\begin{document}

\maketitle

\begin{abstract}
Doctor agents are moving beyond single-turn answer generation toward evolving clinical decision systems that acquire evidence within episodes and reuse experience across episodes. Fixed-input medical QA and many interactive benchmarks provide limited support for evaluating how episode-level decisions interact with cross-episode experience. We introduce \textsc{MedEvoEval}, an executable longitudinal evaluation framework based on action-gated simulated outpatient episodes. Each case is converted into role-specific patient, examination, and manager views; evidence is revealed through valid actions, and structured traces link observations, actions, final outputs, scores, and optional experience write-back. Experiments show that these traces expose process costs hidden by final-answer scoring, characterize how MDT-style consultation reallocates resources, and support analyses of memory maturation, held-out transfer, update-stage response, and backward retention. On 200 held-out cases, the mature-memory condition with sequential write-back reduces interaction turns by 0.915 and improves composite utility by 0.0483 after Holm correction. \textsc{MedEvoEval} provides a concrete basis for evaluating experience-driven improvement, transfer, and retention. The accompanying artifact provides 700 processed episode records with provenance, role-specific schemas, and execution, scoring, and analysis code.
\end{abstract}

\section{Introduction}
\label{sec:introduction}

Doctor agents are increasingly evaluated as clinical decision systems rather than single-turn answer generators. In an outpatient episode, an agent must decide what to ask, which examinations to request, when to seek consultation, and when the available evidence is sufficient to support a diagnosis and management plan. When memory, retrieval, reflection, or other update mechanisms are enabled, later decisions may also depend on prior episodes. The evaluation target therefore shifts from a single medical answer to an evolving clinical decision process.

Current evaluation practice only partially covers this target. Exam-style medical QA benchmarks supply the question or case vignette and score the final answer against a reference~\citep{jin2021medqa,pal2022medmcqa}. These benchmarks do not evaluate action-gated acquisition of patient-specific evidence, action validity, or process-level resource use. Recent benchmarks and simulation environments evaluate interactive diagnosis, simulated consultations, tool use, and EHR-based tasks~\citep{li2024agenthospital,almansoori2025medagentsim,jiang2025medagentbench,nori2025sdbench}. Many of these settings, however, still evaluate individual encounters or fixed benchmark runs. Once prior episodes can affect later decisions through memory or other updates, evaluation must also measure experience-driven behavioral change across episodes.

This evaluation problem has two coupled layers. Within an episode, the evaluator must observe how the agent acquires evidence, uses resources, handles invalid or unavailable actions, and reaches a final diagnosis. Across episodes, the evaluator must measure whether experience improves later behavior, whether the improvement transfers to unseen cases, and whether later updates preserve earlier-stage performance. These layers are linked in doctor-agent evaluation: the trajectory that explains one episode can also become the experience source that influences later episodes.

We introduce \textsc{MedEvoEval}, an executable longitudinal evaluation framework for doctor agents. Each source case is converted into role-specific patient, examination, and manager views. During an outpatient episode, evidence is revealed only through valid actions such as inquiry and examination requests, and every action is recorded in an event trace. After finalization, the manager scores the diagnosis, supporting evidence, and management plan. Optional experience write-back allows later episodes to retrieve compact memory records under controlled conditions, making experience reuse a measurable evaluation condition.

We instantiate this protocol as a runnable E\&D artifact and evaluate it with episode-level and longitudinal studies. The experiments show that traces expose hidden process costs, MDT-style consultation mainly shifts resource allocation, and accumulated memory improves held-out interaction efficiency and composite utility under the reported protocol. The longitudinal setup also measures within-stream changes and earlier-stage retention. This paper makes three contributions.

\textbf{We introduce a framework linking within-episode decisions with cross-episode evaluation.} \textsc{MedEvoEval} specifies information access, valid actions, evidence release, scoring, and experience write-back, supporting comparisons of clinical outcomes, process costs, transfer, and retention.

\textbf{We provide a reusable dataset and evaluation toolkit for doctor agents.} The release includes 700 processed outpatient episode records with role-specific views, source attribution, schemas, a runner, configurations, scoring scripts, process records, and analysis code. Researchers can evaluate compatible doctor models in the supplied environment or adapt their agents to its action and output formats.

\textbf{Our experiments reveal distinct changes in clinical quality and resource use.} Model comparisons and MDT studies show how episode traces expose process costs and consultation changes resource allocation. Memory and held-out studies show gains in interaction efficiency and composite utility, while update studies assess earlier-stage retention.

\section{Related work}
\label{sec:related_work}

\paragraph{Medical QA and fixed-input evaluation.}
Medical QA benchmarks such as MedQA and MedMCQA evaluate exam-style medical knowledge~\citep{jin2021medqa,pal2022medmcqa}, while medical LLM studies such as generalist foundation-model evaluations and Med-PaLM 2 show progress toward expert-level medical question answering~\citep{singhal2025medpalm,nori2023generalist}. More recent resources and studies broaden evaluation toward EHR-like records and sequential diagnosis~\citep{johnson2023mimiciv,jiang2025medagentbench,nori2025sdbench}, while medical LLM work such as PMC-LLaMA illustrates the growing model ecosystem~\citep{wu2024pmcllama}. Exam-style medical QA remains useful for measuring medical knowledge and final-answer quality. Its fixed-input format provides the question context upfront and offers limited information about how an agent obtains patient-specific evidence through clinical actions. Table~\ref{tab:contract_vs_static} contrasts fixed-input medical QA with the information-access, action-validity, resource-use, memory, stability, and auditability controls exposed by \textsc{MedEvoEval}.

This distinction matters for doctor-agent evaluation because two systems with similar final-answer accuracy can follow different clinical paths. One agent may ask targeted questions and request a small number of discriminative examinations, whereas another may rely on broad testing, premature finalization, or unsupported evidence. \textsc{MedEvoEval} evaluates the episode trajectory in addition to the final output, allowing outcome quality to be interpreted together with evidence acquisition and resource use.

\begin{table}[t]
\centering
\renewcommand{\arraystretch}{1.08}
\caption{What \textsc{MedEvoEval} controls beyond fixed-input medical QA.}
\label{tab:contract_vs_static}
\footnotesize
\setlength{\tabcolsep}{3.5pt}
\begin{tabular}{p{0.24\linewidth}p{0.29\linewidth}p{0.36\linewidth}}
\toprule
\rowcolor{TableHeader}
Dimension & Fixed-input QA & \textsc{MedEvoEval} \\
\midrule
Information access & Full vignette given upfront & Evidence revealed by \texttt{ASK} and \texttt{REQUEST\_TEST} \\
Action validity & Not measured & Invalid tests and premature finalization logged \\
Resource use & Invisible & Turns, tests, and invalid-test rate measured \\
Memory effects & Usually absent & No-memory, cold-start, and mature-memory conditions compared \\
Stability & Not evaluated & Held-out performance, update response, and BWT reported \\
Auditability & Final answer only & Structured event logs and retrieved memory cards retained \\
\bottomrule
\end{tabular}
\end{table}

\begin{figure}[!t]
    \centering
    \includegraphics[width=\linewidth]{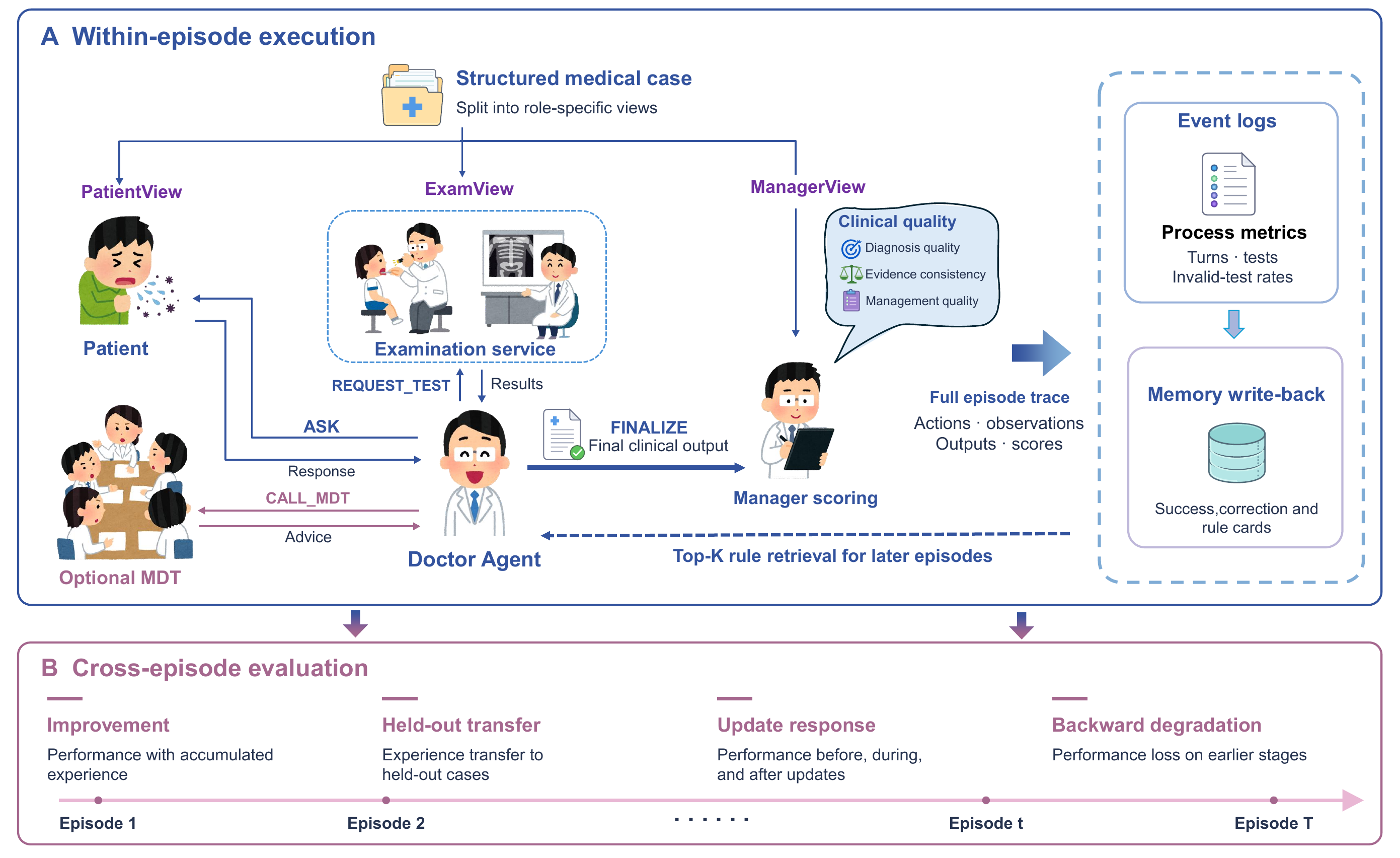}
    \caption{Overview of MedEvoEval. (A) Structured cases are split into patient, examination, and manager views. Doctor agents acquire evidence through inquiry and examination, optionally consult an MDT advisor, and submit final clinical outputs for manager scoring. Event logs record actions, observations, outputs, and scores, supporting process-cost measurement and optional experience write-back with top-$K$ rule retrieval for later episodes. (B) The resulting episode records support analyses of improvement, held-out transfer, update response, and backward degradation.}
    \label{fig:framework_overview}
\end{figure}

\paragraph{Interactive agents, tools, and medical simulation.}
General LLM-agent research has explored reasoning-action loops, tool use, modular tool routing, multi-agent communication, and simulated social behavior~\citep{yao2023react,schick2023toolformer,karpas2022mrkl,shen2023hugginggpt,li2023camel,wu2023autogen,park2023generativeagents,wang2024llmagent}. Medical-agent systems adapt these ideas to diagnosis, consultation, and clinical workflow. MedAgents, MDAgents, MMedAgent, MedAide, AI Hospital, Agent Hospital, MedAgentBench, and MedAgentSim model collaboration, tool use, patients, examinations, EHR environments, or hospital-like interactions~\citep{tang2024medagents,kim2024mdagents,li2024mmedagent,wei2024medaide,fan2025aihospital,li2024agenthospital,jiang2025medagentbench,almansoori2025medagentsim}. MedAgentSim further studies self-evolving multi-agent clinical simulation with doctor, patient, and measurement agents plus experience buffers~\citep{almansoori2025medagentsim}. DxEvolve develops a self-evolving diagnostic agent with interactive evidence acquisition and reusable diagnostic cognition primitives, evaluated on held-out and external cohorts~\citep{ren2026dxevolve}. Our contribution focuses on the evaluation protocol and its reusable records and measurements for comparing outcome quality, process cost, transfer, and retention.

These systems move evaluation closer to clinical workflow by exposing parts of the within-episode decision process. A remaining challenge is to measure how interaction traces and experience mechanisms influence later decisions. \textsc{MedEvoEval} addresses this challenge as an E\&D evaluation artifact: it provides a released processed episode corpus, construction files, role-specific schemas, gated evidence release, structured event logs, reproducible scoring, held-out transfer tests, and update-retention analysis.

This artifact-level control matters for medical-agent evaluation. Prompts, state transitions, information access, action validity, consultant authority, and memory exposure can all change the interpretation of an evaluation result. \textsc{MedEvoEval} makes these components explicit in the benchmark interface, allowing episode execution, scoring, and longitudinal comparisons to be inspected.

\paragraph{Memory, retrieval, and continual evaluation.}
Continual learning studies adaptation under sequential data while preserving earlier capabilities~\citep{lopezpaz2017gradient,parisi2019continual,wang2024clsurvey,wu2024clllms,shi2025clllms}. Retrieval-augmented generation combines language generation with externally retrieved context~\citep{lewis2020retrieval,gao2023ragsurvey,yu2024ragevalsurvey}. Agentic RAG also incorporates reflection into retrieval workflows~\citep{singh2025agenticrag}. Some LLM agents adapt through retrieved reflections or accumulated rules while keeping model parameters fixed~\citep{shinn2023reflexion,yang2023failures}. Medical simulators have begun to show that memory can improve average performance~\citep{li2024agenthospital,almansoori2025medagentsim}. For doctor agents, however, average improvement is only one part of the evaluation problem. A useful protocol should also test whether experience reduces process cost, transfers to unseen cases, remains stable after updates, and avoids backward degradation. \textsc{MedEvoEval} brings these requirements into a single medical-agent evaluation setting by combining controlled evidence access within episodes with longitudinal measurements of experience-driven change across episodes.

\section{\textsc{MedEvoEval}: an executable evaluation protocol}
\label{sec:framework}

\textsc{MedEvoEval} evaluates doctor agents as evolving clinical decision systems through standardized, executable outpatient episodes. As shown in Figure~\ref{fig:framework_overview}, a source case first enters the system as a structured clinical record and is converted into role-specific patient, examination, and manager views. The case is then executed as an action-gated outpatient episode. At runtime, the doctor starts from a brief opening complaint rather than the full patient view, gathers patient facts through inquiry, requests examinations through a controlled test interface, optionally consults an MDT-style advisor, and finally submits a diagnosis, supporting evidence, management plan, and follow-up. After finalization, the manager scores the output, the system records structured event logs, and compact experience records may be written back for later episodes. Across the resulting episode sequence, \textsc{MedEvoEval} analyzes improvement, held-out transfer, update response, and backward degradation.

The protocol turns the two-layer evaluation problem introduced in Section~\ref{sec:introduction} into an executable procedure. Within each episode, it controls what information the doctor can observe and which actions can reveal new evidence. Across episodes, it records trajectories, manager feedback, and optional experience write-back so that later behavior can be evaluated as a function of prior experience. The environment models outpatient first-visit diagnostic episodes because this setting has a clear clinical boundary while preserving the key properties removed by fixed-input QA: partial initial information, active evidence acquisition, explicit resource use, and controlled experience updates.

\paragraph{Design goals.}
The framework is built around role-specific observability, gated evidence release, separate measurement of outcome and process, and longitudinal adaptation analysis. These requirements make unnecessary tests, missed questions, hallucinated findings, memory shortcuts, and backward degradation visible in the evaluation record. \textsc{MedEvoEval} controls what evidence is visible, which actions are valid, how many turns and tests are used, whether memory is available, and how transfer or update stability is measured.

\subsection{Role-specific episode views}
\label{subsec:role_views}

Each case is represented as three role-specific views:
\begin{equation*}
    c_i = (V_i^{\mathrm{patient}}, V_i^{\mathrm{exam}}, V_i^{\mathrm{manager}}).
\end{equation*}
This representation is both a data format and the access-control mechanism that makes each episode executable. The patient view contains demographics, chief complaint, history, and facts that can be revealed through consultation; the examination view maps allowed test names to examination results that remain hidden until explicitly requested; and the manager view contains the reference diagnosis, acceptable aliases, evidence rubrics, and management rubrics used after finalization for scoring and feedback. The doctor observes only the current visible state and never directly accesses unrevealed examination fields, manager-side rubrics, or provenance metadata. The transformation pipeline converts existing source cases into this three-view representation by separating consultation-revealed patient facts from examination-only findings, normalizing test names against a controlled catalog, and constructing manager-side rubrics for diagnosis, evidence, and management. Each transformed episode retains source metadata for attribution and downstream source requirements, and the released construction files preserve an audit path from the source manifest to processed episode fields. Two physicians audited the same 50 source-to-episode conversions; both judged the target diagnosis preserved in all 50 cases. Appendix~\ref{app:conversion_audit} reports the conversion criteria and results, and Appendix~\ref{app:schema} gives the schema and a synthetic sample.

\subsection{Action-gated evidence acquisition}
\label{subsec:action_gated_protocol}

The doctor selects from four structured actions:
\begin{equation*}
\mathcal{A}=\{\texttt{ASK},\texttt{REQUEST\_TEST},\texttt{CALL\_MDT},\texttt{FINALIZE}\}.
\end{equation*}
\texttt{ASK} sends a question to the patient simulator, which answers only from the patient-visible facts. \texttt{REQUEST\_TEST} queries the examination service for a specific catalog item; a result is returned when the requested test is available for the current case, and unavailable or invalid requests are recorded. \texttt{CALL\_MDT} optionally invokes a consultant and is disabled when the experiment isolates single-doctor memory. \texttt{FINALIZE} terminates the episode and requires a final diagnosis, supporting evidence, a management plan, and follow-up.

After each action, the environment updates the observable history and records the event. The next doctor decision is conditioned on the accumulated observable history. Resource limits, including the maximum number of doctor-action attempts and the maximum number of tests, are specified in configuration files. These constraints allow different agents and conditions to be compared under the same visible state, action space, evidence-release rules, and resource budget.

\subsection{Event traces and manager scoring}
\label{subsec:event_trace_scoring}

Each episode produces both a final clinical output and a process trace. The JSONL event log records the turn index, action type, question or requested test, returned-result status, MDT status, termination reason, final outputs, judge scores, and memory write-back status. These fields connect episode execution to later measurement: turn counts and examination counts come from the action trace, invalid-test rates come from test-return status, diagnosis and management scores come from manager-side judging, and memory effects can be inspected against the retrieved cards shown to the doctor.

Manager scoring is applied after \texttt{FINALIZE}. The manager evaluates the submitted diagnosis, supporting evidence, management plan, and follow-up against the manager-side rubric. This process produces normalized diagnosis, evidence, and management scores, together with auxiliary indicators such as strict correctness, pass/fail status, and unsafe or unsupported elements. The trace therefore supports decomposition of a low utility score into weak diagnosis, unsupported evidence, long consultation, excessive examination use, or invalid test requests.

\subsection{Experience write-back and cross-episode evaluation}
\label{subsec:experience_writeback}

The same trace creates the longitudinal link between one episode and later episodes. After scoring, the manager may write compact rule-like cards into success and failure-correction memory banks. Success cards capture reusable \texttt{IF ... THEN ...} patterns, correction cards capture \texttt{IF ... AVOID ...} failure modes, and rule cards capture action conditions such as \texttt{ACTION=ASK}, \texttt{ACTION=REQUEST\_TEST}, \texttt{ACTION=CALL\_MDT}, or \texttt{ACTION=FINALIZE}. Before later decisions, TF--IDF retrieval selects the top-$K$ relevant rule cards for the doctor context ($K=3$ in the main memory experiments). The memory bank grows without an eviction limit. Card-length measurements are reported in Appendix~\ref{app:event_log}; a 100-case frozen-bank comparison with a common card count and truncation policy is reported in Appendix~\ref{app:memory_mechanism}.

We treat retrieve-and-replay memory as a controlled evaluation condition. Memory can be disabled, initialized as empty, or pre-populated from previous stages while the doctor model, patient simulator, test catalog, scoring rubric, and case stream remain fixed. These settings hold the model roles, patient simulation, examination availability, and scoring rules fixed while varying memory access and accumulation. This design also provides the basis for longitudinal comparisons of memory maturation, held-out transfer, update response, and backward retention.

\subsection{Released artifact}
\label{subsec:released_artifact}

The accompanying \textsc{Chinese MedEvoEval-MedQA-700} artifact contains 700 processed episode records. We use 500 records for the longitudinal stream and 200 for held-out evaluation. Cross-sectional and MDT experiments reuse a 100-record subset of the stream; the four-model comparison uses 80 of these records. The stream and held-out sets have no shared case content. The data card records source mappings and repeated content within the stream. The artifact includes role-view records, schemas, the experiment runner, configurations, scoring, sanitized process records, and analysis scripts (Appendix~\ref{app:artifact}). \ArtifactAvailability

\section{Evaluation protocol}
\label{sec:statistical_evaluation}

To address the within-episode and cross-episode questions introduced in Section~\ref{sec:introduction}, we use two levels of measurement. Episode-level outcomes and process costs support cross-sectional comparisons, while longitudinal metrics track experience-driven changes, held-out transfer, and retention after updates. Each episode produces a structured final output and an event log. The final output is used to score diagnosis, evidence, and management, while the event log is used to compute interaction turns, examination use, invalid actions, and memory exposure. Full rubrics, parsing rules, parameter settings, and sensitivity checks are provided in Appendix~\ref{app:scoring}.

\paragraph{Episode-level outcome.}
For episode $i$, the manager assigns three normalized scores: diagnostic quality $S_i^{\mathrm{diag}}$, evidence consistency $S_i^{\mathrm{evi}}$, and management quality $S_i^{\mathrm{plan}}$. The diagnostic score uses a graded rubric, while evidence and management scores compare the final structured output with manager-side rubric points. The total outcome score is
\begin{equation*}
S_i
=
0.5S_i^{\mathrm{diag}}
+
0.3S_i^{\mathrm{evi}}
+
0.2S_i^{\mathrm{plan}}.
\end{equation*}
We also report strict diagnostic accuracy and pass rate. Strict accuracy is the fraction of episodes with $S_i^{\mathrm{diag}}=1$, as implemented in the scoring code. Pass rate counts episodes whose total outcome score reaches the predefined threshold, $S_i\geq 0.7$.

\paragraph{Process cost and utility.}
Process metrics are computed from event logs. Let $T_i$ count \texttt{ASK}, \texttt{REQUEST\_TEST}, and failed doctor-action attempts. \texttt{FINALIZE} and \texttt{CALL\_MDT} are excluded from $T_i$; MDT calls are recorded separately. Let $E_i$ count requested examinations and $R_i$ be the fraction with no nonempty returned result, set to zero when $E_i=0$. The execution budget is $B=10$ doctor-action attempts. The utility uses a fixed turn-cost divisor $C_T=B=10$ and an examination divisor $E_{\max}=3$:
\begin{equation*}
U_i
=
0.6S_i
-
0.2(T_i/C_T)
-
0.1(E_i/E_{\max})
-
0.1R_i.
\end{equation*}
Utility represents benchmark performance under a fixed outcome-cost weighting. We report $U_i$ together with its components so that outcome quality, turns, examinations, and invalid-test behavior remain separately inspectable.

\paragraph{Experience reuse and held-out transfer.}
For longitudinal analysis, a case stream is partitioned into ordered stages $W_1,\ldots,W_M$. For any episode-level metric $Z_i$, the stage mean is
\begin{equation*}
\bar{Z}_m
=
\frac{1}{|W_m|}
\sum_{i\in W_m}
Z_i.
\end{equation*}
Held-out evaluation uses 200 cases outside the 500-record longitudinal stream, in a common order. $H_0$ disables retrieval and write-back. $H_3$ starts from the stream's mature memory and retains retrieval and sequential write-back during evaluation. This comparison measures held-out behavior under these two experience conditions. For any metric $Z$,
\begin{equation*}
\mathrm{ET}(Z)
=
\bar{Z}_{H_3}
-
\bar{Z}_{H_0}.
\end{equation*}
Positive transfer in outcome or utility indicates better held-out performance, while negative transfer in turns or examinations indicates lower process cost.

\paragraph{Update response and retention.}
For streams with an explicit update interval, we summarize performance before, during, and after the interval. For metric $Z$, update response is measured by $\Delta_{\mathrm{upd}}(Z)=\bar{Z}_{\mathrm{upd}}-\bar{Z}_{\mathrm{pre}}$, and subsequent change by $\Delta_{\mathrm{post}}(Z)=\bar{Z}_{\mathrm{post}}-\bar{Z}_{\mathrm{upd}}$.
To measure whether later updates preserve earlier-stage performance, we evaluate earlier stages after the system has been updated through later stages. Let $A_{k,j}^{(Z)}$ be the value of metric $Z$ on stage $j$ after the system has been updated through stage $k$. For higher-is-better metrics, backward transfer is
\begin{equation*}
\mathrm{BWT}(Z)
=
\frac{1}{K-1}
\sum_{j=1}^{K-1}
\left(
A_{K,j}^{(Z)}
-
A_{j,j}^{(Z)}
\right).
\end{equation*}
Negative values indicate backward degradation. We report BWT primarily for $S$ and $U$ to keep the sign convention consistent.

\paragraph{Statistical comparisons.}
Most comparisons use paired case sets. For continuous or ordinal metrics, including total outcome score, utility, turns, examination count, invalid-test rate, and diagnostic quality, we report paired differences and use Wilcoxon signed-rank tests. For paired binary outcomes such as strict diagnostic correctness and pass/fail status, we use exact McNemar tests. Mean estimates and paired differences are accompanied by 95\% bootstrap confidence intervals computed by resampling case records while preserving the pairing. We apply Holm correction within each stated comparison family at $\alpha=0.05$. The held-out family contains all seven reported endpoints. Appendix~\ref{app:statistical_tests} lists family membership, effect sizes, confidence intervals, and both raw and adjusted $p$-values. Reported significance refers to the adjusted values; the intervals describe the individual effect estimates.

\section{Experiments}
\label{sec:experiments}

The experiments are staged to answer progressively broader evaluation questions. Section~\ref{subsec:exp_model_discriminability} first tests whether episode traces expose differences hidden by answer-only scoring. Section~\ref{subsec:exp_mdt_resource} then treats MDT consultation as a controlled process intervention with the primary doctor fixed. Section~\ref{subsec:exp_memory_transfer} finally asks whether experience reuse matures on the stream, transfers to held-out cases, and remains stable after an update.

Unless otherwise specified, episodes use an action budget of 10, $E_{\max}=3$, local runtime seed 7, and the same scoring code. Paired conditions share case sets and order. Table~\ref{tab:settings} summarizes the principal settings; Appendix~\ref{app:splits_repro} records generation settings and compute accounting.

\begin{table}[t]
\centering
\renewcommand{\arraystretch}{1.08}
\caption{Principal experiment settings. The patient is Qwen-Turbo and the manager is Kimi-K2.5. Memory conditions use top-$K=3$ retrieval; enabled memory is writable unless explicitly marked frozen.}
\label{tab:settings}
\small
\setlength{\tabcolsep}{4pt}
\begin{tabular}{llrl}
\toprule
\rowcolor{TableHeader}
Study & Primary doctor & Cases & Intervention \\
\midrule
Model comparison & Four models in Table~\ref{tab:static_dynamic} & 80 & Memory and MDT off \\
MDT & Qwen-Turbo & 100 & Off / Qwen-Turbo or Kimi-K2.5 consultant \\
Longitudinal & Qwen-Turbo & 500 & E0--E3 memory conditions \\
Held-out & Qwen-Turbo & 200 & H0 off / H3 mature, writable \\
Strong-primary MDT & GLM-5 & 100 & Off / GLM-5 consultant \\
Interface control & Four models & 80 & Raw / normalized; 640 episodes \\
\bottomrule
\end{tabular}
\end{table}

\begin{figure}[t]
    \centering
    \includegraphics[width=0.95\textwidth]{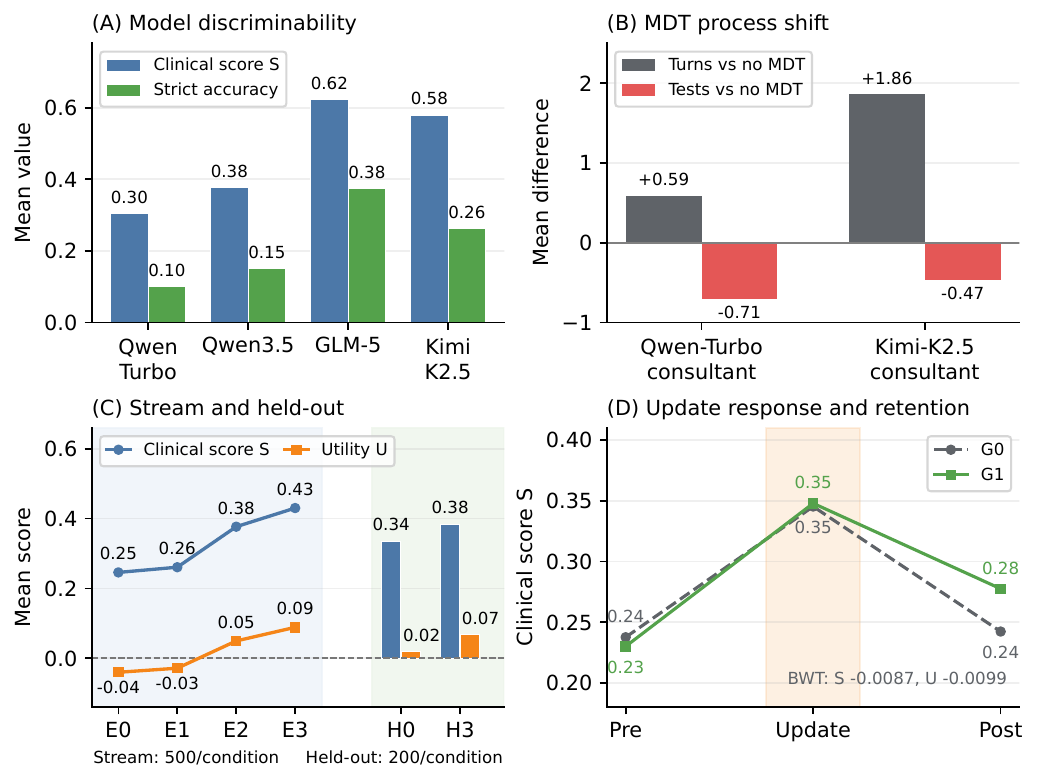}
    \caption{\textsc{MedEvoEval} experiment overview. (A) Clinical score and strict accuracy distinguish the four doctor models. (B) MDT changes in turns and examinations relative to no MDT. (C) Clinical score (blue) and utility (orange): lines compare E0--E3 stream conditions (500 records each), and bars compare H0/H3 held-out conditions (200 records each); the shaded regions distinguish the two case sets. (D) Pre-update, update, and post-update scores, with backward-transfer estimates.}
    \label{fig:experimental_overview}
\end{figure}

\subsection{What does the episode trace add beyond final-answer scoring?}
\label{subsec:exp_model_discriminability}

As shown in Figure~\ref{fig:experimental_overview}(A), this experiment asks what final-answer scoring leaves unobserved. We compare four API-accessible doctor models on the same 80 outpatient episodes with memory and MDT disabled. Table~\ref{tab:static_dynamic} reports each outcome once and adds the utility and process measures available from the episode trace.

\begin{table}[t]
\centering
\renewcommand{\arraystretch}{1.08}
\caption{Outcome and episode-trace metrics on 80 shared episodes. Strict accuracy and $S$ also constitute the answer-only projection. Blue highlights the highest strict accuracy, $S$, and $U$; purple marks the process costs discussed in the text.}
\label{tab:static_dynamic}
\small
\setlength{\tabcolsep}{4.2pt}
\begin{tabular}{lcccccc}
\toprule
\rowcolor{TableHeader}
& \multicolumn{3}{c}{Outcome and utility $\uparrow$} & \multicolumn{3}{c}{Process cost} \\
\cmidrule(lr){2-4}\cmidrule(lr){5-7}
Model & Strict & $S$ & $U$ & Turns & Tests & Invalid \\
\midrule
Qwen-Turbo & 0.100 & 0.304 & 0.014 & 5.713 & 1.388 & 0.077 \\
Qwen3.5-122B-A10B & 0.150 & 0.378 & 0.036 & \transferfocus{7.975} & 0.900 & 0.010 \\
\rowcolor{TableFocusBlue}
\textbf{GLM-5} & \scorebest{0.375} & \scorebest{0.622} & \scorebest{0.205} & 4.463 & 2.225 & 0.050 \\
Kimi-K2.5 & 0.263 & 0.579 & 0.128 & 5.725 & \transferfocus{2.575} & \transferfocus{0.192} \\
\bottomrule
\end{tabular}
\end{table}

\paragraph{Episode traces expose process cost.}
The trace view reveals differences that answer-only scoring does not show. GLM-5 obtains the highest total score and utility in this subset. Kimi-K2.5 achieves strong outcome quality, but it uses more examinations and has a higher invalid-test rate. Qwen3.5-122B-A10B requests fewer examinations, but it uses more turns and obtains lower outcome quality. These patterns decompose final-answer quality, examination behavior, invalid actions, and utility. Appendix~\ref{app:model_comparison_supp} reports confidence intervals and paired deltas. Across the 320 original model-comparison episodes, 82 (25.6\%) contain at least one recorded error; this is an episode-level incidence. A separate 640-episode paired raw/normalized-interface control improves Qwen3.5's score and utility after correction while preserving the four-model ordering in both arms (Appendix~\ref{app:interface_control}). The four-model score ranking is preserved for 60 of 66 score-weight settings (Appendix~\ref{app:weight_sensitivity}).

\subsection{Does MDT change diagnostic quality or resource allocation?}
\label{subsec:exp_mdt_resource}

As shown in Figure~\ref{fig:experimental_overview}(B), this experiment uses MDT-style consultation as a process intervention. We keep Qwen-Turbo as the primary doctor and compare no MDT with Qwen-Turbo and Kimi-K2.5 consultants on 100 shared case records. Full paired results are reported in Appendix~\ref{app:mdt_full_results}.

\paragraph{MDT changes resource allocation more than diagnostic accuracy.}
In the Qwen-Turbo setting, MDT-style consultation mainly changes resource allocation. Examination use decreases from 1.37 tests per episode to 0.66 with a Qwen-Turbo consultant and 0.90 with a Kimi-K2.5 consultant. Kimi consultation increases interaction turns from 5.70 to 7.56. Neither consultant yields a Holm-corrected significant gain in clinical score, strict accuracy, or utility. A complementary 100-case experiment holds a stronger GLM-5 primary doctor fixed and uses GLM-5 as consultant; MDT increases interaction turns by 1.33 and reduces examinations by 0.56 (both Holm-adjusted $p<0.001$), without a corrected-significant outcome gain. Its full paired estimates are reported alongside the Qwen-Turbo results in Appendix~\ref{app:mdt_full_results}.

\subsection{Does experience reuse improve, transfer, and retain?}
\label{subsec:exp_memory_transfer}

Figure~\ref{fig:experimental_overview}(C,D) and Table~\ref{tab:memory_stability} separate repeated-stream performance, held-out performance, and earlier-stage retention.

We evaluate a single-doctor agent without MDT on the same 500-case stream, split into ten 50-case stages. E0 disables memory; E1 starts with empty writable memory; E2 and E3 repeat the stream with the accumulated bank from the preceding pass. These comparisons therefore combine repeated exposure and memory accumulation. On 200 held-out cases, H0 disables retrieval and write-back, while H3 starts from the mature bank and continues sequential write-back. BWT measures earlier-stage retention after later adaptation.

\begin{table}[t]
\centering
\renewcommand{\arraystretch}{1.08}
\caption{Longitudinal memory, transfer, and update-retention evidence. Blue highlights mature stream memory (E3); bold purple values mark the two corrected-significant held-out improvements. Longitudinal summaries and held-out paired estimates are in Appendix~\ref{app:memory_transfer_supp}; update-retention diagnostics are in Appendix~\ref{app:update_retention}.}
\label{tab:memory_stability}
\footnotesize
\setlength{\tabcolsep}{3.5pt}
\begin{tabular}{llccccc}
\toprule
\rowcolor{TableHeader}
Study & Setting & $N$ & $S$ & $U$ & Strict/Pass & Turns \\
\midrule
Main stream & E0 & 500 & 0.246 & -0.040 & 0.092 / 0.056 & 6.134 \\
Main stream & E1 & 500 & 0.261 & -0.029 & 0.136 / 0.096 & 6.324 \\
Main stream & E2 & 500 & 0.377 & 0.050 & 0.276 / 0.244 & 5.696 \\
\rowcolor{TableFocusBlue}
Main stream & \textbf{E3} & 500 & \scorebest{0.430} & \scorebest{0.089} & \scorebest{0.342} / \scorebest{0.284} & \scorebest{5.380} \\
\midrule
Held-out & H0 & 200 & 0.335 & 0.019 & 0.260 / 0.180 & 6.645 \\
\rowcolor{TableFocusPurple}
Held-out & \textbf{H3} & 200 & 0.385 & \transferfocus{0.068} & 0.295 / 0.255 & \transferfocus{5.730} \\
\midrule
Update & G0 pre/update/post & 500 & 0.238/0.345/0.242 & -0.043/0.025/-0.037 & -- & -- \\
Update & G1 pre/update/post & 500 & 0.230/0.348/0.277 & -0.055/0.046/-0.013 & -- & -- \\
BWT & Final memory agent & 9 stages & -0.0087 & -0.0099 & -- & -- \\
\bottomrule
\end{tabular}
\end{table}

\paragraph{Experience memory shows a maturity pattern.}
E1 has small outcome and utility gains and uses more turns than E0. By E2 and E3, outcome score, utility, strict accuracy, and pass rate increase while turns decline. From E0 to E3, total score rises from 0.246 to 0.430, strict accuracy from 0.092 to 0.342, and turns fall from 6.134 to 5.380; Appendix~\ref{app:memory_transfer_supp} gives the full longitudinal summary.

\paragraph{Held-out transfer tests reusable experience.}
On the 200 paired held-out cases, with Holm correction across seven endpoints, H3 reduces interaction turns by 0.915 (95\% CI $[-1.290,-0.540]$, adjusted $p<0.001$) and increases composite utility by 0.0483 (95\% CI $[0.0185,0.0784]$, adjusted $p=0.017$). Appendix~\ref{app:memory_transfer_supp} reports paired differences, confidence intervals, and test results for all endpoints.

\paragraph{Update response and retention measure stability.}
\label{subsec:exp_update_retention}
The update stream contains middle stages with shifted symptom combinations, examination dependencies, or case-expression patterns. Table~\ref{tab:memory_stability} summarizes performance before, during, and after the update interval; its BWT row reports final-stage retention over nine earlier stages, with supporting diagnostics in Appendix~\ref{app:update_retention}.

G0 disables memory retrieval, whereas G1 enables retrieval and online write-back with top-$K=3$. Both settings follow the same 500-record sequence, with stages T1--T4, T5--T6, and T7--T10 forming the pre-update, update, and post-update windows. The update stages improve performance during the shifted interval for both G0 and G1, while post-update performance differs across settings. BWT gives the complementary retention view: $\mathrm{BWT}_{S}=-0.0087$ and $\mathrm{BWT}_{U}=-0.0099$, which are small negative point estimates of backward transfer. Reporting update response and BWT together separates current-stage adaptation from earlier-stage retention.

\paragraph{Scoring validation.}
Two physicians independently scored the same 80 outputs while blinded to each other and to automatic scores. They assessed diagnosis, evidence, and management on outputs sampled from model-comparison, stream-memory, and held-out conditions. Binary diagnostic agreement has $\kappa=0.824$ between physicians. On this audit sample, both physicians and the automatic scorer agree on the direction in 21 of 24 group-mean comparisons (Appendix~\ref{app:llm_sim_audit}). Alternative-judge and diagnosis-matching analyses cover 2,520 recorded outputs; the same-judge repeat covers 520. The alternative judges and repeat scorer preserve the four-model score ordering; Appendix~\ref{app:judge_robustness} specifies the covered subsets.

\section{Discussion, limitations, and ethics}
\label{sec:discussion_limitations}

The experiments support evaluating doctor agents at both levels. Within episodes, similar final-answer scores can hide different evidence-acquisition paths, examination use, invalid actions, and unsupported-evidence risks. Across episodes, apparent improvement must be interpreted with transfer and retention, because memory and update mechanisms can change later behavior.

For E\&D, the main value of \textsc{MedEvoEval} is that the mechanism is executable and inspectable. The released schemas, runner, configurations, event logs, scoring scripts, and analysis code show how evidence is revealed, actions are recorded, scores are computed, and longitudinal comparisons are derived. The resulting audit trail can be reweighted, checked for invalid actions, rerun on the corpus, and used to compare memory policies under the same runner.

\textsc{MedEvoEval} binds information access, action validity, resource use, memory exposure, scoring, held-out transfer, and post-update retention into one protocol, making improvements easier to interpret as changes in clinical output, process cost, experience reuse, or stability.

The framework has important limitations. The episodes are simulated and standardized: patient responses come from structured views, examination results are pre-specified, and manager scores depend on rubrics and automatic judging. Appendix~\ref{app:llm_sim_audit} reports an independent two-physician audit of manager-side scoring. Case transformation may introduce artifacts in wording, examinations, disease distribution, or rubrics; composite scores encode value choices; and the repeated-stream gains combine exposure and memory accumulation. The held-out H3 policy also updates its memory during evaluation. The controlled update event is a retention probe rather than a full robustness benchmark, and the baselines cover four API-accessible models rather than open-source families or broad scale sweeps.

 \textsc{MedEvoEval} scores are controlled evaluation signals, not deployment evidence. Autonomous diagnosis or patient-facing use would require expert validation, institutional review, privacy safeguards, regulatory compliance, and human oversight.

Future work should broaden sources, add larger physician-adjudicated subsets, evaluate open-source and differently scaled models, test alternative MDT policies, and stress-test memory under stronger shifts. Separating factual medical retrieval from experiential process memory is also important because they raise different governance questions.

\section{Conclusion}
\label{sec:conclusion}

We presented \textsc{MedEvoEval}, an executable longitudinal evaluation framework for doctor agents. The framework evaluates doctor agents as evolving clinical decision systems by combining action-gated outpatient episodes, event-level traces, controlled experience reuse, held-out transfer tests, and update-retention analysis. The experiments show that the protocol exposes process costs hidden by final-answer scoring, characterizes how MDT-style consultation reallocates resources, and evaluates experience-driven improvement, held-out transfer, and backward retention. On 200 held-out cases, mature writable memory reduces interaction turns and improves composite utility after Holm correction. These results support process-aware and longitudinal evaluation as a practical basis for studying continually evolving doctor agents.

\Needspace{9\baselineskip}
\begin{ack}
The author thanks the two physicians for their time and clinical expertise in reviewing the source-to-episode conversions and independently assessing the doctor-agent outputs.

\textbf{Funding.} This work received no external funding.

\textbf{Competing interests.} \CompetingInterestsStatement
\end{ack}

\bibliographystyle{plainnat}
\bibliography{references}

\appendix
\raggedbottom
\renewcommand{\topfraction}{0.92}
\renewcommand{\bottomfraction}{0.85}
\renewcommand{\textfraction}{0.06}
\renewcommand{\floatpagefraction}{0.72}
\setlength{\textfloatsep}{12pt plus 2pt minus 2pt}
\setlength{\floatsep}{10pt plus 2pt minus 2pt}
\section{Artifact inventory and responsible use}
\label{app:artifact}

The accompanying artifact contains the research runner, a separate lightweight demonstration, scoring and analysis code, configurations, schemas, 700 Chinese MedQA-derived case records, sanitized process trajectories, manifests, data documentation, and Croissant metadata. Frozen results and analysis scripts support offline verification without model API calls. The demonstration uses simplified patient, consultant, and judge components; the research runtime and recorded experiments are supplied separately.

Table~\ref{tab:app_release_boundary} lists the accompanying materials and their uses.

\begin{table}[!htbp]
\centering
\caption{Contents of the accompanying evaluation artifact.}
\label{tab:app_release_boundary}
\small
\begin{tabular}{p{0.30\linewidth}p{0.58\linewidth}}
\toprule
Component & Purpose \\
\midrule
Case records and manifests & 700 records with stable release identifiers, partition membership, source provenance, and content-duplicate annotations. \\
Process derivatives & 3,103 complete-episode trajectories and 3,386 single-step decision samples. \\
Research runtime and demonstration & Inspect the action-gated experiment implementation and run a separate lightweight example. \\
Configurations and schemas & Document model roles, budgets, memory settings, and structured case, action, score, and event fields. \\
Frozen results and reproduction scripts & Recompute paired statistics, corrected $p$-values, table data, and vector figures offline. \\
Documentation and metadata & Describe the data, licenses, provenance, usage boundaries, and Croissant metadata for the prepared release. \\
\bottomrule
\end{tabular}
\end{table}

\paragraph{Sources and licenses.}
\label{app:asset_licenses}
The 700 case records are derived from Chinese MedQA~\citep{jin2021medqa}; the release preserves its MIT license and copyright notice. Project-added code, schemas, documentation, and process derivatives retain the supplied CC BY-NC-SA 4.0 license. Model inference uses API services under the Alibaba Cloud Model Studio agreement and the applicable third-party model terms. The release's asset notice lists the source links, component licenses, and service terms.

\paragraph{Training-oriented derivatives.}
The 3,103 complete-episode trajectories are accompanied by 3,386 single-step decision samples extracted from high-quality episodes. Each step records the current context, previous questions and tests, available actions, a target action and output, and the episode score. These files support research on action policies. Low-scoring and failed episodes are excluded from the step derivative.

\paragraph{Responsible use.}
\textsc{MedEvoEval} is a simulation-based evaluation framework for research use. Its episodes and scores do not establish clinical readiness, medical-device certification, or safety for real-world deployment. The data and logs support benchmark inspection and reproducibility, not patient care.

\section{Executable episode protocol details}
\label{app:protocol_details}

This appendix groups the executable protocol details used by the runner: the role-specific case schema, runtime prompt templates, the doctor structured-output template, and a compact memory-record example.

\subsection{Role-specific case schema}
\label{app:schema}

The implementation defines an episode case as a structured object with patient, examination, and manager views. The doctor-facing state is derived from this object and exposes only an opening instruction, revealed history, returned tests, optional consultant notes, optional retrieved memory cards, and the allowed test catalog. Figure~\ref{fig:app_role_views} summarizes the role-specific visibility constraints, and the synthetic sample illustrates the case format.

\begin{figure}[!htbp]
\centering
\begin{tikzpicture}[
  font=\small,
  box/.style={draw, rounded corners=2pt, align=center, minimum width=3.2cm, minimum height=1cm},
  note/.style={draw, rounded corners=2pt, align=left, minimum width=4.0cm, inner sep=4pt},
  arrow/.style={-Latex, thick}
]
\node[box, fill=gray!10] (case) at (0,0) {Structured case};
\node[box, fill=blue!8] (patient) at (-4,-2) {PatientView\\patient simulator};
\node[box, fill=green!8] (exam) at (0,-2) {ExamView\\exam service};
\node[box, fill=orange!10] (manager) at (4,-2) {ManagerView\\manager judge};
\node[note, fill=purple!7] (doctor) at (0,-4.3) {Doctor sees:\\$\bullet$ opening complaint\\$\bullet$ revealed history\\$\bullet$ returned tests\\$\bullet$ consultant notes (optional)\\$\bullet$ retrieved memory (optional)};
\node[note, fill=red!6] (hidden) at (6.0,-4.3) {Doctor does \emph{not} see:\\$\bullet$ hidden patient facts\\$\bullet$ unrevealed tests\\$\bullet$ manager answer key\\$\bullet$ scoring rubric internals};

\draw[arrow] (case) -- (patient);
\draw[arrow] (case) -- (exam);
\draw[arrow] (case) -- (manager);
\draw[arrow] (patient) -- (doctor);
\draw[arrow] (exam) -- (doctor);
\draw[arrow] (manager) -- ++(0,-0.7) -| (hidden.north);
\end{tikzpicture}
\caption{Role-specific case views and visibility constraints. Patient and examination information is revealed only through actions, while manager-side labels and rubrics remain hidden until evaluation.}
\label{fig:app_role_views}
\end{figure}

\paragraph{Synthetic sample.}
The following example is synthetic and only illustrates the format.

\begin{codebox}{Schema B.1: Synthetic case example}
\begin{lstlisting}[language=json]
{
  "case_id": "synthetic_demo_001",
  "source": {"type": "synthetic", "raw_case_id": "demo"},
  "patient_view": {
    "demographics": {"age": 46, "sex": "female"},
    "chief_complaint": "intermittent chest discomfort for two days",
    "history_facts": [
      "pain worsens with exertion",
      "no fever",
      "family history of premature coronary disease"
    ]
  },
  "exam_view": {
    "tests": {
      "electrocardiogram": "nonspecific ST-T changes",
      "troponin": "mildly elevated"
    },
    "test_catalog": ["electrocardiogram", "troponin"]
  },
  "manager_view": {
    "ground_truth": {
      "diagnosis": {
        "display_name": "acute coronary syndrome",
        "acceptable_names": ["ACS"]
      },
      "evidence_rubric": {
        "key_points": ["exertional chest pain", "troponin elevation"]
      },
      "management_rubric": {
        "key_points": ["urgent cardiology evaluation"]
      }
    }
  }
}
\end{lstlisting}
\end{codebox}

\FloatBarrier
\subsection{Prompt templates}
\label{app:prompts}

The boxes below are faithful English translations or labeled excerpts of the Chinese prompts in the released research engine (\texttt{clinical/prompts.py}); the model-facing prompts and responses use Chinese. Braced placeholders stand for runtime values. Prompt~B.1 shows the branch with examination requests and consultation enabled; other branches remove unavailable actions. The complete source retains all branches, turn-specific reminders, and repair prompts.

\begin{promptbox}{Prompt B.1: Attending-doctor system prompt (translation)}
\begin{lstlisting}
You are an outpatient attending doctor.
All outward-facing content must be in Chinese.
Output only one JSON object.
The only allowed actions are ASK, REQUEST_TEST, CALL_MDT, FINALIZE.
REQUEST_TEST requests physical, auxiliary, or imaging examinations;
test_name must be selected exactly from the available catalog.
CALL_MDT invokes one consultant for diagnostic review, not a vote.
Consultant advice is for reference; the final decision remains yours.
CALL_MDT is allowed only if (question_count >= 4 or test_count >= 1)
and no consultant has yet been called. At most one call per case.
Test names are matched after trimming whitespace, ignoring case;
do not rewrite the meaning of a catalog item.
For FINALIZE, provide diagnosis, evidence, plan, and followup.
The FINALIZE diagnosis array must contain one clear, short diagnosis
phrase, not multiple candidate diagnoses.
Use only information already released to the doctor; do not invent
examination results.
Output fields must match the template below; do not add fields:
{doctor_json_template}
Available examination catalog:
{available_test_catalog}
Current context:
{context_prompt}
\end{lstlisting}
\end{promptbox}

The template in Prompt~B.1 is reproduced in Appendix~\ref{app:structured_outputs}. The context contains the full dialogue, released examination results, retrieved rule cards, consultant summary, and runtime constraints. The separate user prompt supplies the current turn, question and examination counts, returned results, and requested examinations. Its reminders require a nonempty question containing at most two questions for \texttt{ASK}, no repeated questions, an empty question field for other actions, and nonempty final-output arrays with exactly one diagnosis for \texttt{FINALIZE}.

\begin{promptbox}{Prompt B.2: Patient system and user prompts (translation)}
\begin{lstlisting}
System:
You are the patient in an outpatient simulation.
All outward-facing content must be in Chinese.
Answer only with information the patient knows, can feel, or can recall.
Patient profile: {patient_profile}

User:
Doctor's question: {question}
Return only one JSON object, containing only the field "utterance".
\end{lstlisting}
\end{promptbox}

Here, \texttt{patient\_profile} serializes demographics, chief complaint, history of present illness, past medical history, social history, and review of systems. Examination results and the manager's scoring rubric are not part of that profile.

\begin{codebox}{Protocol B.3: Deterministic examination lookup (pseudocode)}
\begin{lstlisting}
key = lower(strip(requested_test_name))
if the examination budget is exceeded:
    return returned=false, match_type="budget_limited"
if key is absent from the normalized allowed catalog:
    return returned=false, match_type="not_allowed"
if key is absent from the normalized case examination map:
    return returned=false, match_type="not_in_case"
return test_name=stored_name, returned=true,
       result_text=stored_result, match_type="exact"
\end{lstlisting}
\end{codebox}

The examination service performs this deterministic lookup without an LLM call, alias expansion, or paraphrasing. Its response fields are \texttt{test\_name}, \texttt{returned}, \texttt{result\_text}, \texttt{evidence\_tags}, and \texttt{match\_type}. The orchestrator handles the budget check; \texttt{TestService} handles catalog and case lookup. Only successful results are appended to the released-results list.

\begin{promptbox}{Prompt B.4: Consultant system prompt (translated excerpt)}
\begin{lstlisting}
You are an outpatient consultant reviewing the attending doctor's
current diagnostic reasoning. All outward-facing content is Chinese.
You cannot request examinations or submit a final diagnosis directly
to the environment.
Output only one JSON object matching the template; add no fields.
advice_action can only be ASK or FINALIZE.
For ASK, recommended_question is nonempty and contains at most two
specific questions; recommended_diagnosis is an empty array.
For FINALIZE, recommended_diagnosis has length one;
recommended_question is an empty array.
confidence is a number from 0 to 1.
Do not invent examination results that have not appeared.
Output template:
{consultant_json_template}
Current context:
{context_prompt}
\end{lstlisting}
\end{promptbox}

The consultant template has the fields \texttt{advice\_action}, \texttt{utterance}, \texttt{consult\_opinion}, \texttt{working\_diagnoses}, \texttt{key\_evidence\_for}, \texttt{key\_evidence\_against}, \texttt{missing\_points}, \texttt{recommended\_question}, \texttt{recommended\_diagnosis}, \texttt{confidence}, and \texttt{rationale}. Its user prompt supplies the interaction counts and available results and requests the differential, supporting and opposing evidence, missing information, and a next-question or finalization recommendation.

\FloatBarrier
\subsection{Structured output templates}
\label{app:structured_outputs}

\begin{codebox}{Template B.1: Doctor structured output}
\begin{lstlisting}[language=json]
{
  "action": "ASK|REQUEST_TEST|CALL_MDT|FINALIZE",
  "utterance": "string",
  "question": "string",
  "test_name": "string",
  "rationale": ["string"],
  "working_diagnoses": ["string"],
  "needs_mdt": false,
  "diagnosis": ["string"],
  "evidence": ["string"],
  "plan": ["string"],
  "followup": ["string"]
}
\end{lstlisting}
\end{codebox}

\FloatBarrier
\subsection{Example memory record}
\label{app:event_log}

The following synthetic memory-card example shows the memory-record format without including any third-party-derived case content.

\begin{codebox}{Log B.2: Synthetic memory-card examples}
\begin{lstlisting}[language=json]
[
  {
    "case_id": "synthetic_demo_001",
    "visit_index": 0,
    "memory_type": "success",
    "experience": "Exertional chest discomfort plus biomarker elevation supported early cardiac workup.",
    "rule_text": "IF chest discomfort worsens with exertion THEN ask cardiac-risk questions and prioritize ECG/troponin.",
    "outcome": "accepted diagnosis and supported management plan",
    "search_text": "exertional chest discomfort cardiac risk ECG troponin acute coronary syndrome"
  }
]
\end{lstlisting}
\end{codebox}

\paragraph{Retrieval and memory size.}
The research runtime retrieves rule cards using TF--IDF similarity and presents at most $K_{\mathrm{ret}}=3$ rule cards per doctor decision, unless the retrieval-scale ablation changes this setting. The memory bank stores successful experiences, corrective experiences, and rule cards, with continued write-back and no fixed capacity limit or eviction. At the end of the three memory-enabled passes of 500 records each, the bank contains 1,500, 3,000, and 4,500 records, respectively. At the end of E3, the 4,500 records comprise 312 successful experiences, 1,188 corrective experiences, and 3,000 rule cards.

In the relevant-retrieval arm of the frozen-bank control, a prompt-level audit covers 594 doctor decision turns, each containing three rule cards. Their combined length has a median of 329 characters, approximately 164 tokens under an empirical calibration using provider-reported token usage. Memory content accounts for 15.9\% of the total characters across the audited prompts. A separate card-length audit uses a 3,204-rule-card snapshot accumulated after E3 and subsequent demonstration and historical held-out runs. Individual card lengths have a median of 128 characters, an interquartile range of 98--154, a 95th percentile of 195, and a maximum of 332. The implementation truncates cards at 80 whitespace-separated units.

\section{Scoring rubric and metric details}
\label{app:scoring}

This appendix expands the scoring rules and metric definitions used in Section~\ref{sec:statistical_evaluation}. The main paper reports the core quantities; this appendix provides the rubric levels, default parameters, process definitions, longitudinal notation, and sensitivity settings.

\FloatBarrier
\subsection{Outcome rubric}

Diagnostic quality is graded on five levels:
\[
S_i^{\mathrm{diag}}\in\{1.00,0.75,0.50,0.25,0.00\}.
\]
The levels correspond to an exact or clinically equivalent diagnosis, a mostly correct diagnosis with minor deviation, a partially correct diagnosis with moderate omission or bias, a limited but non-dangerous error, and a dangerous or clearly wrong diagnosis. Two physicians independently assessed the scoring behavior on the output sample described in Appendix~\ref{app:llm_sim_audit}.

Evidence consistency compares the final supporting evidence with manager-side key evidence points. Let $K_i^{\mathrm{evi}}$ be the number of key evidence points, $m_i^{\mathrm{evi}}$ the number of correctly matched points, and $h_i^{\mathrm{evi}}$ the number of contradicted, hallucinated, or unsupported evidence points. We define
\begin{equation*}
S_i^{\mathrm{evi}}
=
\max\left(
0,
\frac{
m_i^{\mathrm{evi}}
-
\lambda_{\mathrm{evi}}h_i^{\mathrm{evi}}
}{
K_i^{\mathrm{evi}}
}
\right).
\end{equation*}

Management quality is scored in the same way. Let $K_i^{\mathrm{plan}}$ be the number of expected management points, $m_i^{\mathrm{plan}}$ the number of matched points, and $h_i^{\mathrm{plan}}$ the number of unsafe or clearly wrong management points. We define
\begin{equation*}
S_i^{\mathrm{plan}}
=
\max\left(
0,
\frac{
m_i^{\mathrm{plan}}
-
\mu_{\mathrm{plan}}h_i^{\mathrm{plan}}
}{
K_i^{\mathrm{plan}}
}
\right).
\end{equation*}

The primary strict-accuracy endpoint requires the implemented diagnosis grade to be exact:
\begin{equation*}
A_i^{\mathrm{strict}}=\mathbf{1}\{S_i^{\mathrm{diag}}=1\}.
\end{equation*}
The implemented scoring procedure first matches the normalized diagnosis against the reference diagnoses, then applies the management-strategy rule, which lowers the diagnosis grade when the management plan has a major deviation. Strict accuracy uses the resulting diagnosis grade. Appendix~\ref{app:judge_robustness} examines the contribution of this second step through a diagnosis-matching check.
Pass rate uses the composite score threshold:
\begin{equation*}
P_i=\mathbf{1}\{S_i\geq\tau_{\mathrm{pass}}\}.
\end{equation*}

\FloatBarrier
\subsection{Default parameters}

Default scoring and utility parameters are listed in Table~\ref{tab:app_default_params}.

\begin{table}[!htbp]
\centering
\caption{Default scoring and utility parameters.}
\label{tab:app_default_params}
\small
\begin{tabular}{lc}
\toprule
Parameter & Value \\
\midrule
Outcome weights $(w_{\mathrm{diag}},w_{\mathrm{evi}},w_{\mathrm{plan}})$ & $(0.5,0.3,0.2)$ \\
Evidence penalty $\lambda_{\mathrm{evi}}$ & 0.5 \\
Management penalty $\mu_{\mathrm{plan}}$ & 1.0 \\
Pass threshold $\tau_{\mathrm{pass}}$ & 0.7 \\
Utility weights $(w_1,w_2,w_3,w_4)$ & $(0.6,0.2,0.1,0.1)$ \\
Execution action budget and test limit & $B=10$, $E_{\max}=3$ \\
Turn-cost reference scale & $C_T=B=10$ \\
\bottomrule
\end{tabular}
\end{table}

With these parameters, the composite outcome score is
\begin{equation*}
S_i
=
w_{\mathrm{diag}}S_i^{\mathrm{diag}}
+
w_{\mathrm{evi}}S_i^{\mathrm{evi}}
+
w_{\mathrm{plan}}S_i^{\mathrm{plan}},
\end{equation*}
and the utility score is
\begin{equation*}
U_i
=
w_1S_i
-
w_2\frac{T_i}{C_T}
-
w_3\frac{E_i}{E_{\max}}
-
w_4R_i.
\end{equation*}

\FloatBarrier
\subsection{Process metrics}

The event log records each action before finalization. The interaction count is
\begin{equation*}
T_i
=
\#\texttt{ASK}
+
\#\texttt{REQUEST\_TEST}
+
N_i^{\mathrm{error}},
\end{equation*}
where $N_i^{\mathrm{error}}$ counts failed doctor-action attempts. Both \texttt{CALL\_MDT} and \texttt{FINALIZE} are excluded from this reported metric. A failed action consumes the execution budget and increases $T_i$ by one; it has no direct deduction from diagnosis, evidence, or plan scores. An MDT action also consumes the execution budget, and its invocation frequency is reported separately. The turn cost is normalized by the configured action budget, $C_T=B=10$. The examination count is
\begin{equation*}
E_i=\#\texttt{REQUEST\_TEST}.
\end{equation*}
Let $I_i$ be the number of examination requests for which the service returns no result or only blank result text. The invalid-test rate is
\begin{equation*}
R_i
=
\begin{cases}
I_i/E_i, & E_i>0,\\
0, & E_i=0.
\end{cases}
\end{equation*}

\FloatBarrier
\subsection{Longitudinal metrics}

For a staged stream $W_1,\ldots,W_M$, the stage mean of metric $Z$ is
\begin{equation*}
\bar{Z}_m
=
\frac{1}{|W_m|}
\sum_{i\in W_m}
Z_i.
\end{equation*}
The early and late averages used for learning gain are
\begin{equation*}
\bar{Z}_{\mathrm{early}}
=
\frac{1}{|\mathcal{M}_{\mathrm{early}}|}
\sum_{m\in\mathcal{M}_{\mathrm{early}}}
\bar{Z}_m,
\qquad
\bar{Z}_{\mathrm{late}}
=
\frac{1}{|\mathcal{M}_{\mathrm{late}}|}
\sum_{m\in\mathcal{M}_{\mathrm{late}}}
\bar{Z}_m.
\end{equation*}
Thus,
\begin{equation*}
\mathrm{LG}(Z)=\bar{Z}_{\mathrm{late}}-\bar{Z}_{\mathrm{early}}.
\end{equation*}

For held-out evaluation, $H_0$ and $H_3$ use the same ordered set $\mathcal{H}$ of 200 case records (scenario IDs 500--699; release IDs \texttt{case\_501}--\texttt{case\_700}), excluded from the 500-record experience stream. $H_0$ runs without memory; $H_3$ starts from the mature stream-derived memory and permits online write-back during the held-out sequence. Consequently, later $H_3$ episodes can retrieve experience from earlier held-out episodes. The paired contrast evaluates this sequential memory-use condition on unseen, same-source cases. We summarize its average difference as
\begin{equation*}
\mathrm{ET}(Z)
=
\frac{1}{|\mathcal{H}|}
\sum_{i\in\mathcal{H}}
\left(
Z_i^{(H_3)}
-
Z_i^{(H_0)}
\right).
\end{equation*}
The direction of interpretation depends on the metric. Positive values are favorable for higher-is-better metrics such as $S$, $U$, strict accuracy, and pass rate. Negative values are favorable for cost metrics such as $T$ and $E$.

For retention analysis, $A_{k,j}^{(Z)}$ denotes performance on stage $j$ after the system has been updated through stage $k$. For higher-is-better metrics,
\begin{equation*}
\mathrm{BWT}(Z)
=
\frac{1}{K-1}
\sum_{j=1}^{K-1}
\left(
A_{K,j}^{(Z)}
-
A_{j,j}^{(Z)}
\right).
\end{equation*}
Negative BWT indicates backward degradation.

\FloatBarrier
\subsection{Update-stage response}

When a stream contains an explicit update interval, let $\mathcal{M}_{\mathrm{pre}}$, $\mathcal{M}_{\mathrm{upd}}$, and $\mathcal{M}_{\mathrm{post}}$ denote stages before, during, and after the update. For metric $Z$,
\begin{equation*}
\Delta_{\mathrm{upd}}(Z)
=
\bar{Z}_{\mathrm{upd}}
-
\bar{Z}_{\mathrm{pre}},
\qquad
\Delta_{\mathrm{post}}(Z)
=
\bar{Z}_{\mathrm{post}}
-
\bar{Z}_{\mathrm{upd}}.
\end{equation*}
If $\bar{Z}_{\mathrm{upd}}<\bar{Z}_{\mathrm{pre}}$ for a higher-is-better metric, recovery is summarized as
\begin{equation*}
\rho_{\mathrm{recovery}}(Z)
=
\frac{
\bar{Z}_{\mathrm{post}}
-
\bar{Z}_{\mathrm{upd}}
}{
\bar{Z}_{\mathrm{pre}}
-
\bar{Z}_{\mathrm{upd}}
}.
\end{equation*}
This ratio is only interpreted when the update interval produces a drop.

\FloatBarrier
\subsection{Statistical tests}
\label{app:statistical_tests}

For paired conditions $A$ and $B$, we report the average record-level difference $d_i^{(Z)}=Z_i^{(B)}-Z_i^{(A)}$. Continuous or ordinal endpoints use two-sided paired Wilcoxon signed-rank tests; binary endpoints use two-sided exact McNemar tests. Effect sizes, paired bootstrap 95\% confidence intervals, raw $p$-values, and Holm-adjusted $p$-values are reported together. Bootstrap resampling preserves the condition pairing. The final held-out analysis uses 100,000 resamples per endpoint (seeds 507--513); the remaining paired intervals use 20,000 resamples with seed 7. The percentile intervals summarize mean differences, and the Wilcoxon tests evaluate paired ranks. For H0/H3, resampling draws case pairs from the observed sequence, during which H3 continues online memory write-back.

Holm correction is applied within explicitly listed scientific comparison families. Each family groups endpoints and, where applicable, contrasts addressing the same question. The full final test manifest records the family, comparison, endpoint, sample size, test, effect, and raw and corrected $p$-values. We interpret corrected results alongside their effect sizes and intervals, and treat small update, retention, retrieval-scale, and non-significant outcome contrasts as exploratory estimates.

The three-level retrieval-scale comparison uses paired Friedman tests for continuous endpoints and Cochran $Q$ tests for binary endpoints, with six endpoints in one Holm family. Table~\ref{tab:app_holm_families} enumerates the final 18 families; the code package provides all 141 tests, including non-significant results, in \href{\CodeURL/blob/main/results/final_holm_tests.csv}{\nolinkurl{results/final_holm_tests.csv}}.

For stage sequences, Mann--Kendall tests summarize monotonic trends, and cumulative sums of centered stage means (CUSUM) describe stage-level shifts. Pre-update, update, and post-update means summarize the response across the specified windows. These are descriptive diagnostics based on ten stage observations.

\FloatBarrier
\subsection{Sensitivity}
\label{app:weight_sensitivity}
We vary the nonnegative outcome weights $(w_d,w_e,w_p)$ over the full simplex with spacing 0.1 and $w_d+w_e+w_p=1$, yielding 66 settings, including 36 with all three weights positive. Each score is recomputed from the same logged diagnosis, evidence, and plan components. Utility is then recomputed with its default cost weights and fixed turn reference $C_T=10$. Strict diagnostic accuracy is unchanged by outcome weighting.

The four-model total-score ordering GLM-5 $>$ Kimi-K2.5 $>$ Qwen3.5 $>$ Qwen-Turbo holds at 60/66 settings and 35/36 interior settings. The six exceptions swap only GLM-5 and Kimi-K2.5 and assign at least 0.8 weight to plan quality. E3 exceeds E0 at all 66 settings; the stricter E0 $<$ E1 $<$ E2 $<$ E3 ordering holds at 58/66. The final held-out H3--H0 differences in both total score and utility are positive at all 66 settings. The sweep summarizes mean differences and ranking changes across weight settings.

The default $(0.5,0.3,0.2)$, equal-thirds, $(0.4,0.4,0.2)$, and $(0.6,0.2,0.2)$ alternatives also preserve the four-model total-score ordering. Table~\ref{tab:app_named_weights} gives their scores and mean memory contrasts.

\begin{table}[!htbp]
\centering
\caption{Named outcome-weight alternatives, including equal thirds (shown rounded in the weight column). Model columns show total score; the final columns show mean differences.}
\label{tab:app_named_weights}
\scriptsize
\setlength{\tabcolsep}{3pt}
\begin{tabular}{llrrrrrrr}
\toprule
Setting & $(w_d,w_e,w_p)$ & QT & Q3 & GLM & Kimi & $\Delta E$ & $\Delta H_S$ & $\Delta H_U$ \\
\midrule
default & $(0.50,0.30,0.20)$ & 0.304 & 0.378 & 0.622 & 0.579 & 0.184 & 0.049 & 0.048 \\
balanced & $(0.40,0.40,0.20)$ & 0.303 & 0.389 & 0.626 & 0.586 & 0.163 & 0.044 & 0.045 \\
diagnosis heavy & $(0.60,0.20,0.20)$ & 0.305 & 0.367 & 0.618 & 0.573 & 0.206 & 0.054 & 0.051 \\
equal thirds & $(0.33,0.33,0.33)$ & 0.320 & 0.397 & 0.649 & 0.618 & 0.164 & 0.054 & 0.051 \\
\bottomrule
\end{tabular}
\end{table}

\begin{figure}[!htbp]
\centering
\includegraphics[width=\linewidth]{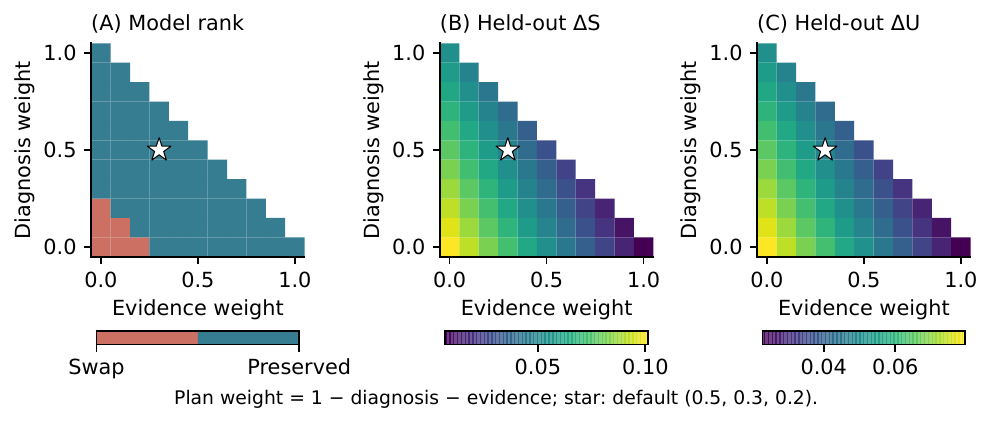}
\caption{All 66 outcome-weight settings. (A) Preservation of the default four-model score ranking. (B) Held-out H3--H0 total-score difference. (C) Held-out H3--H0 utility difference with turn reference 10.}
\label{fig:app_weight_sensitivity}
\end{figure}

Figure~\ref{fig:app_weight_sensitivity} shows model-ranking preservation and held-out differences across all 66 weight settings.

The complete 66-setting numerical grid is provided in \path{results/weight_sensitivity_grid_66.csv} in the code package.

Cost-weight sensitivity is distinct from outcome-weight sensitivity. Table~\ref{tab:app_utility_sensitivity} changes the four utility weights while retaining the default outcome score and the same recorded episodes. All settings use turns/10 and tests/3. GLM-5 ranks first under all five settings; the remaining order depends on the chosen resource penalty.

\begin{table}[!htbp]
\centering
\caption{Utility under five cost-weight settings on the same 80-record model comparison. Weights are ordered as (score, turns, tests, invalid-test rate), with divisors 10 and 3. Bold marks the highest utility in each row.}
\label{tab:app_utility_sensitivity}
\scriptsize
\setlength{\tabcolsep}{4pt}
\begin{tabular}{llrrrr}
\toprule
Setting & Weights & QT & Q3 & GLM-5 & Kimi \\
\midrule
Default & $(0.6,0.2,0.1,0.1)$ & 0.014 & 0.036 & \textbf{0.205} & 0.128 \\
Outcome heavy & $(0.8,0.1,0.1,0.0)$ & 0.140 & 0.193 & \textbf{0.379} & 0.320 \\
Turn cost heavy & $(0.4,0.4,0.1,0.1)$ & -0.161 & -0.199 & \textbf{-0.009} & -0.102 \\
Test cost heavy & $(0.5,0.1,0.3,0.1)$ & -0.052 & 0.018 & \textbf{0.039} & -0.044 \\
Invalid-test heavy & $(0.5,0.1,0.1,0.3)$ & 0.025 & 0.076 & \textbf{0.177} & 0.089 \\
\bottomrule
\end{tabular}
\end{table}

\FloatBarrier
\section{Reproducibility notes, additional results, and ablations}
\label{app:stat_results}

This appendix consolidates the reproduction notes and supplementary experimental results for Section~\ref{sec:experiments}. The results are organized by their role in the paper: model-comparison supplements, MDT consultation supplements, longitudinal memory and held-out transfer, update and retention diagnostics, and memory ablations. Exact split membership, commands, and configuration files are provided in the accompanying artifact.

\FloatBarrier
\subsection{Reproducibility and operational notes}
\label{app:splits_repro}

\paragraph{Execution and resources.}
The experiments use API-based inference, with no local model training or fine-tuning. The default local runtime seed is 7, the configured action budget is $B=10$, and the examination limit is three. The utility turn-cost divisor is ten as specified in Appendix~\ref{app:scoring}. Model-role assignments, sampled record identifiers, pairings, and memory settings are retained in the accompanying configuration and result manifests. The resource ledger reports the available API-call, prompt-token, completion-token, and timing records by experiment.
Generation settings are listed in Table~\ref{tab:app_generation_settings}, and resource use in Table~\ref{tab:app_resource_usage}.
\begin{table}[!htbp]
\centering
\caption{Generation settings. Output caps are requested maximum token counts. QT denotes Qwen-Turbo.}
\label{tab:app_generation_settings}
\small
\setlength{\tabcolsep}{5pt}
\begin{tabular}{p{0.23\linewidth}p{0.68\linewidth}}
\toprule
Setting & Value \\
\midrule
Primary doctor & \texttt{qwen-turbo}, \texttt{qwen3.5-122b-a10b}, \texttt{glm-5}, or \texttt{kimi-k2.5}, according to the experiment. \\
Consultant & QT study: \texttt{qwen-turbo} or \texttt{kimi-k2.5}; strong-primary study: \texttt{glm-5}. \\
Other roles & Patient: \texttt{qwen-turbo}; manager: \texttt{kimi-k2.5}; alternative judges: \texttt{deepseek-v3.2} and \texttt{qwen-max}. \\
Sampling & Temperature 0.05; provider defaults for \texttt{top\_p} and API sampling seed. Local Python seed: $7+\mathrm{scenario\_id}$. \\
Doctor/patient/consultant caps & 1,200 / 500 / 500. \\
Manager caps & Main experiments and held-out extension: extraction 1,200, learning 1,800, other manager requests 1,200. Strong-primary MDT and interface-control profile: extraction 2,400, learning 3,600, other manager requests 2,400. \\
Thinking request & Enabled for the manager and disabled for the doctor, patient, and single consultant. \\
\bottomrule
\end{tabular}
\end{table}
The released runner provides both manager-token profiles; the default reproduction profile uses the main-experiment caps.

The project required approximately $1.06\times10^5$ API calls, including development runs and supplementary experiments. This estimate combines 80,975 calls in the original research ledger, approximately 23,000 supplementary calls, and 1,749 calls for the Kimi-consultant arm. Table~\ref{tab:app_resource_usage} summarizes available per-experiment usage; the accompanying resource ledger retains the detailed counters.

\begin{table}[!htbp]
\centering
\caption{API resource summary. Input/output tokens are in millions; times are elapsed minutes. M: measured counters; E: estimated; P: measured partial-run coverage; M/E: recorded calls and estimated tokens. A dash denotes an unrecorded quantity.}
\label{tab:app_resource_usage}
\scriptsize
\setlength{\tabcolsep}{4pt}
\begin{tabular}{lrrrrl}
\toprule
Experiment / recorded scope & Calls & Input & Output & Min. & Basis \\
\midrule
E0--E3 stream run inventory & 27,524 & 19.605 & 3.352 & -- & M/E \\
Interface intervention (640 episodes) & 6,416 & 4.203 & 6.904 & -- & M \\
DeepSeek rejudge (2,520 outputs) & 2,520 & 1.366 & 0.123 & 6.9 & M \\
Qwen-Max resumed segment (867 outputs) & 869 & 0.510 & 0.041 & 2.7 & P \\
Same-judge retest (520 outputs) & 520 & 0.297 & 0.028 & 2.1 & M \\
Frozen-bank control (300 episodes) & 3,624 & 2.512 & 0.279 & -- & M \\
Kimi-consultant arm (100 episodes) & 1,749 & 1.463 & 0.223 & -- & M \\
Strong-primary MDT (200 episodes) & $\approx$2,000 & -- & -- & -- & E \\
\bottomrule
\end{tabular}
\end{table}

\paragraph{Operational parsing and interaction errors.}
Malformed or unusable doctor-action attempts are logged as \texttt{llm\_error\_turns}. They consume the action budget and count toward the reported turns and utility penalty. Invalid examination requests are tracked separately. In the four-model comparison, 82 of 320 episodes (25.6\%) contain at least one logged doctor-action error.

Table~\ref{tab:app_operational_noise} reports errors by model, and Table~\ref{tab:app_cross_study_noise} summarizes errors across experiments.

\begin{table}[!htbp]
\centering
\caption{Interface errors in the original four-model comparison. Every model uses the same 80 case records.}
\label{tab:app_operational_noise}
\small
\begin{tabular}{lrrr}
\toprule
Model & Episodes with errors & Incidence & Error turns \\
\midrule
Qwen-Turbo & 0/80 & 0.00\% & 0 \\
Qwen3.5-122B-A10B & 71/80 & 88.75\% & 276 \\
GLM-5 & 10/80 & 12.50\% & 11 \\
Kimi-K2.5 & 1/80 & 1.25\% & 4 \\
\midrule
All models & 82/320 & 25.6\% & 291 \\
\bottomrule
\end{tabular}
\end{table}

\begin{table}[!htbp]
\centering
\caption{Parsing and examination errors across the retained final studies. Errors count episodes with at least one logged doctor-action error; invalid tests count requests without a nonempty returned result.}
\label{tab:app_cross_study_noise}
\scriptsize
\setlength{\tabcolsep}{3pt}
\begin{tabular}{lrrrrr}
\toprule
Study & Episodes & Error episodes & Error turns & Invalid tests & Mean $R$ \\
\midrule
Four-model comparison & 320 & 82 (25.6\%) & 291 & 75 & 0.082 \\
Qwen-primary MDT & 300 & 3 (1.00\%) & 4 & 38 & 0.072 \\
Memory stream E0--E3 & 2000 & 101 (5.05\%) & 299 & 802 & 0.167 \\
Held-out H0/H3 & 400 & 16 (4.00\%) & 51 & 38 & 0.043 \\
Update G0/G1 & 1000 & 52 (5.20\%) & 156 & 313 & 0.128 \\
Strong-primary MDT & 200 & 52 (26.00\%) & 71 & 39 & 0.084 \\
\bottomrule
\end{tabular}
\end{table}

\paragraph{Paired interface normalization.}
\label{app:interface_normalization}
\label{app:interface_control}
To assess the effects of interface errors on model performance and ranking, we compare raw and normalized interfaces across four models on the same 80 case records, completing 640 episodes. The deterministic normalizer repairs JSON formatting, action aliases, and uniquely mappable public test names while preserving medical content and access permissions for examination information, with no additional LLM retries. Normalization reduces Qwen3.5's any-error episode rate from 86.25\% to 5.00\%, increases total score by 0.122 and utility by 0.103, with both gains passing Holm correction; the four-model rankings in total score, utility, and strict accuracy remain unchanged across the two interface conditions. Full paired results are reported in Table~\ref{tab:app_interface_results}.

\begingroup
\scriptsize
\setlength{\tabcolsep}{3pt}
\begin{longtable}{p{0.18\linewidth}p{0.17\linewidth}rrrrrr}
\caption{Raw and normalized interface results on 80 paired records per model. Intervals concern normalized minus raw; Holm adjustment uses the outcome, process, or manipulation-check family identified in the test manifest.}\label{tab:app_interface_results}\\
\toprule
Model & Endpoint & Raw & Normalized & $\Delta$ & 95\% interval & Raw $p$ & Holm $p$ \\
\midrule
\endfirsthead
\toprule
Model & Endpoint & Raw & Normalized & $\Delta$ & 95\% interval & Raw $p$ & Holm $p$ \\
\midrule
\endhead
\bottomrule
\endfoot
Qwen-Turbo & Total score & 0.271 & 0.298 & +0.027 & [-0.017, 0.070] & 0.077 & 0.766 \\
Qwen-Turbo & Utility & -0.019 & -0.008 & +0.010 & [-0.026, 0.045] & 0.225 & 1.000 \\
Qwen-Turbo & Strict accuracy & 0.113 & 0.113 & 0.000 & [-0.050, 0.050] & 1.000 & 1.000 \\
Qwen-Turbo & Turns & 6.162 & 6.312 & +0.150 & [-0.362, 0.675] & 0.690 & 1.000 \\
Qwen-Turbo & Tests & 1.438 & 1.475 & +0.037 & [-0.150, 0.225] & 0.664 & 1.000 \\
Qwen-Turbo & Error turns & 0.125 & 0.025 & -0.100 & [-0.263, 0.013] & 0.197 & 0.790 \\
Qwen-Turbo & Any-error episode & 0.037 & 0.025 & -0.013 & [-0.062, 0.025] & 1.000 & 1.000 \\
\midrule
Qwen3.5 & Total score & 0.339 & 0.461 & \textbf{+0.122} & [0.060, 0.186] & $9.43\times10^{-4}$ & 0.010 \\
Qwen3.5 & Utility & 0.010 & 0.113 & \textbf{+0.103} & [0.062, 0.145] & $2.14\times10^{-5}$ & $2.57\times10^{-4}$ \\
Qwen3.5 & Strict accuracy & 0.163 & 0.200 & +0.037 & [-0.050, 0.125] & 0.581 & 1.000 \\
Qwen3.5 & Turns & 7.812 & 4.700 & -3.112 & [-3.600, -2.600] & $6.46\times10^{-13}$ & $5.17\times10^{-12}$ \\
Qwen3.5 & Tests & 0.988 & 1.825 & +0.838 & [0.600, 1.075] & $3.21\times10^{-8}$ & $2.25\times10^{-7}$ \\
Qwen3.5 & Error turns & 3.212 & 0.050 & -3.163 & [-3.650, -2.675] & $4.37\times10^{-13}$ & $3.06\times10^{-12}$ \\
Qwen3.5 & Any-error episode & \textbf{0.8625} & \textbf{0.0500} & -0.812 & [-0.888, -0.725] & $5.42\times10^{-20}$ & $4.34\times10^{-19}$ \\
\midrule
GLM-5 & Total score & 0.610 & 0.597 & -0.013 & [-0.052, 0.023] & 0.716 & 1.000 \\
GLM-5 & Utility & 0.193 & 0.186 & -0.007 & [-0.033, 0.018] & 0.766 & 1.000 \\
GLM-5 & Strict accuracy & 0.388 & 0.362 & -0.025 & [-0.100, 0.050] & 0.754 & 1.000 \\
GLM-5 & Turns & 4.650 & 4.588 & -0.062 & [-0.338, 0.212] & 0.575 & 1.000 \\
GLM-5 & Tests & 2.237 & 2.225 & -0.013 & [-0.138, 0.100] & 0.817 & 1.000 \\
GLM-5 & Error turns & 0.250 & 0.113 & -0.138 & [-0.287, 0.000] & 0.064 & 0.385 \\
GLM-5 & Any-error episode & 0.200 & 0.100 & -0.100 & [-0.212, 0.000] & 0.115 & 0.577 \\
\midrule
Kimi-K2.5 & Total score & 0.594 & 0.566 & -0.027 & [-0.076, 0.020] & 0.350 & 1.000 \\
Kimi-K2.5 & Utility & 0.129 & 0.117 & -0.013 & [-0.047, 0.021] & 0.621 & 1.000 \\
Kimi-K2.5 & Strict accuracy & 0.200 & 0.212 & +0.013 & [-0.062, 0.087] & 1.000 & 1.000 \\
Kimi-K2.5 & Turns & 6.062 & 5.875 & -0.188 & [-0.588, 0.212] & 0.321 & 1.000 \\
Kimi-K2.5 & Tests & 2.550 & 2.612 & +0.062 & [-0.062, 0.200] & 0.394 & 1.000 \\
Kimi-K2.5 & Error turns & 0.087 & 0.000 & -0.087 & [-0.263, 0.000] & 0.317 & 0.952 \\
Kimi-K2.5 & Any-error episode & 0.013 & 0.000 & -0.013 & [-0.037, 0.000] & 1.000 & 1.000 \\
\end{longtable}
\endgroup

\FloatBarrier
\subsection{Model-comparison supplements}
\label{app:model_comparison_supp}

\begin{figure}[!htbp]
\centering
\includegraphics[width=\linewidth]{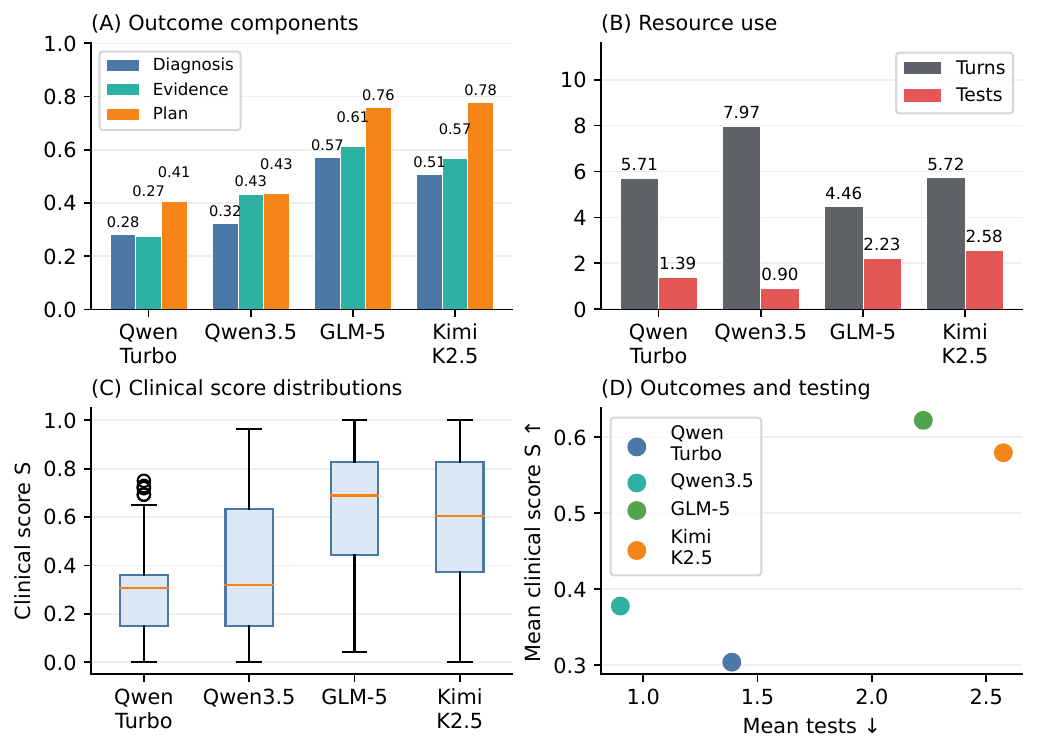}
\caption{Model-comparison evidence on 80 shared episodes. (A) Outcome components separate diagnosis, evidence, and plan. (B) Process cost compares interaction turns and requested tests. (C) Case-level score distributions: the line marks the median, the box spans the 25th--75th percentiles, whiskers extend to the most extreme observations within 1.5 interquartile ranges of the box, and points beyond the whiskers are outliers. (D) Mean total score is plotted against mean examination use for each model; higher positions indicate higher scores, and positions farther left indicate fewer tests.}
\label{fig:app_model_comparison}
\end{figure}

Table~\ref{tab:app_model_pairwise} reports paired differences relative to Qwen-Turbo, and Table~\ref{tab:app_static_dynamic_ci} gives means and intervals. Table~\ref{tab:app_full_vignette_static} compares static complete-input and dynamic episode conditions; Table~\ref{tab:app_exp1_full} and Figure~\ref{fig:app_model_comparison} decompose model outcomes and process behavior.

\begin{table}[!htbp]
\centering
\caption{Paired model contrasts against Qwen-Turbo on the same 80 records. The nine contrasts form one Holm family. Intervals concern paired mean differences.}
\label{tab:app_model_pairwise}
\scriptsize
\setlength{\tabcolsep}{3pt}
\begin{tabular}{llrrrr}
\toprule
Model & Endpoint & $\Delta$ & 95\% interval & Raw $p$ & Holm $p$ \\
\midrule
Qwen3.5 & Total score & +0.074 & [0.005, 0.141] & 0.078 & 0.235 \\
Qwen3.5 & Utility & +0.022 & [-0.021, 0.065] & 0.466 & 0.848 \\
Qwen3.5 & Strict accuracy & +0.050 & [-0.037, 0.138] & 0.424 & 0.848 \\
\midrule
Kimi-K2.5 & Total score & +0.275 & [0.213, 0.339] & $3.03\times10^{-10}$ & $2.12\times10^{-9}$ \\
Kimi-K2.5 & Utility & +0.114 & [0.071, 0.157] & $9.14\times10^{-6}$ & $5.48\times10^{-5}$ \\
Kimi-K2.5 & Strict accuracy & +0.163 & [0.062, 0.263] & 0.004 & 0.018 \\
\midrule
GLM-5 & Total score & +0.318 & [0.259, 0.377] & $1.78\times10^{-12}$ & $1.61\times10^{-11}$ \\
GLM-5 & Utility & +0.191 & [0.153, 0.229] & $8.93\times10^{-12}$ & $7.15\times10^{-11}$ \\
GLM-5 & Strict accuracy & +0.275 & [0.163, 0.388] & $2.74\times10^{-5}$ & $1.37\times10^{-4}$ \\
\bottomrule
\end{tabular}
\end{table}

\begin{table}[!htbp]
\centering
\caption{Model-comparison outcome and process summary with 95\% percentile bootstrap intervals (20,000 resamples; seed 7). Each episode contributes one final outcome score; its trace supplies utility and invalid-test behavior.}
\label{tab:app_static_dynamic_ci}
\scriptsize
\setlength{\tabcolsep}{3pt}
\begin{tabular}{lcccccc}
\toprule
Model & Strict & $S$ & $S$ interval & $U$ & $U$ interval & Invalid \\
\midrule
Qwen-Turbo & 0.100 & 0.304 & [0.261, 0.348] & 0.014 & [-0.015, 0.045] & 0.077 \\
Qwen3.5-122B-A10B & 0.150 & 0.378 & [0.319, 0.438] & 0.036 & [-0.002, 0.076] & 0.010 \\
\rowcolor{TableFocusBlue}
\textbf{GLM-5} & \textbf{0.375} & \textbf{0.622} & [0.567, 0.676] & \textbf{0.205} & [0.169, 0.240] & 0.050 \\
Kimi-K2.5 & 0.263 & 0.579 & [0.519, 0.638] & 0.128 & [0.085, 0.171] & 0.192 \\
\bottomrule
\end{tabular}
\end{table}

To compare final-answer performance with complete information against episodes that acquire information incrementally, we evaluate Qwen-Turbo and Kimi-K2.5 on the same 80 cases under full-vignette static QA and dynamic episode conditions (Table~\ref{tab:app_full_vignette_static}).

\begin{table}[!htbp]
\centering
\caption{Full-vignette static QA results and paired dynamic episode comparison on the 80-case subset for two models. Static evaluation gives the complete case vignette in a single prompt and excludes manager-only fields, gold diagnoses, aliases, and rubrics.}
\label{tab:app_full_vignette_static}
\scriptsize
\setlength{\tabcolsep}{3.2pt}
\begin{tabular}{lc>{\columncolor{TableFocusBlue}}c>{\columncolor{TableFocusBlue}}c>{\columncolor{TableFocusPurple}}c>{\columncolor{TableFocusPurple}}ccccc}
\toprule
Model & $N$ & \textbf{Static strict} & \textbf{Dyn. strict} & \textbf{Static $S$} & \textbf{Dyn. $S$} & Static accept. & Static $S_d$ & Static $S_e$ & Static $S_p$ \\
\midrule
Qwen-Turbo & 80 & 0.288 & 0.100 & 0.585 & 0.304 & 0.388 & 0.472 & 0.682 & 0.722 \\
Kimi-K2.5 & 80 & 0.125 & 0.263 & 0.726 & 0.579 & 0.625 & 0.581 & 0.836 & 0.922 \\
\bottomrule
\end{tabular}
\end{table}

The static condition changes both information availability and interaction demands: the complete vignette is supplied up front, whereas a dynamic episode requires evidence acquisition through actions. We report this as a descriptive comparison. Complete-input final-answer performance and interactive episode performance differ in both magnitude and direction: Qwen-Turbo improves under full-vignette input, whereas Kimi-K2.5 has a higher static composite score but lower static strict diagnostic accuracy than in the dynamic episode condition. Static QA therefore helps contextualize final-answer ability, while episode traces remain necessary for measuring information acquisition, examination cost, invalid actions, and utility.

\begin{table}[!htbp]
\centering
\caption{Full model-comparison summary on 80 shared episodes.}
\label{tab:app_exp1_full}
\small
\begin{tabular}{lcccccccc}
\toprule
Model & Strict & Accept. & $S_d$ & $S_e$ & $S_p$ & $S$ & Turns & Tests \\
\midrule
Qwen-Turbo & 0.100 & 0.175 & 0.281 & 0.274 & 0.406 & 0.304 & 5.713 & 1.388 \\
Qwen3.5-122B-A10B & 0.150 & 0.275 & 0.322 & 0.433 & 0.434 & 0.378 & 7.975 & 0.900 \\
GLM-5 & 0.375 & 0.525 & 0.572 & 0.614 & 0.759 & 0.622 & 4.463 & 2.225 \\
Kimi-K2.5 & 0.263 & 0.488 & 0.506 & 0.569 & 0.778 & 0.579 & 5.725 & 2.575 \\
\bottomrule
\end{tabular}
\end{table}

\subsection{MDT consultation supplements}
\label{app:mdt_full_results}

\begin{figure}[!htbp]
\centering
\includegraphics[width=\linewidth]{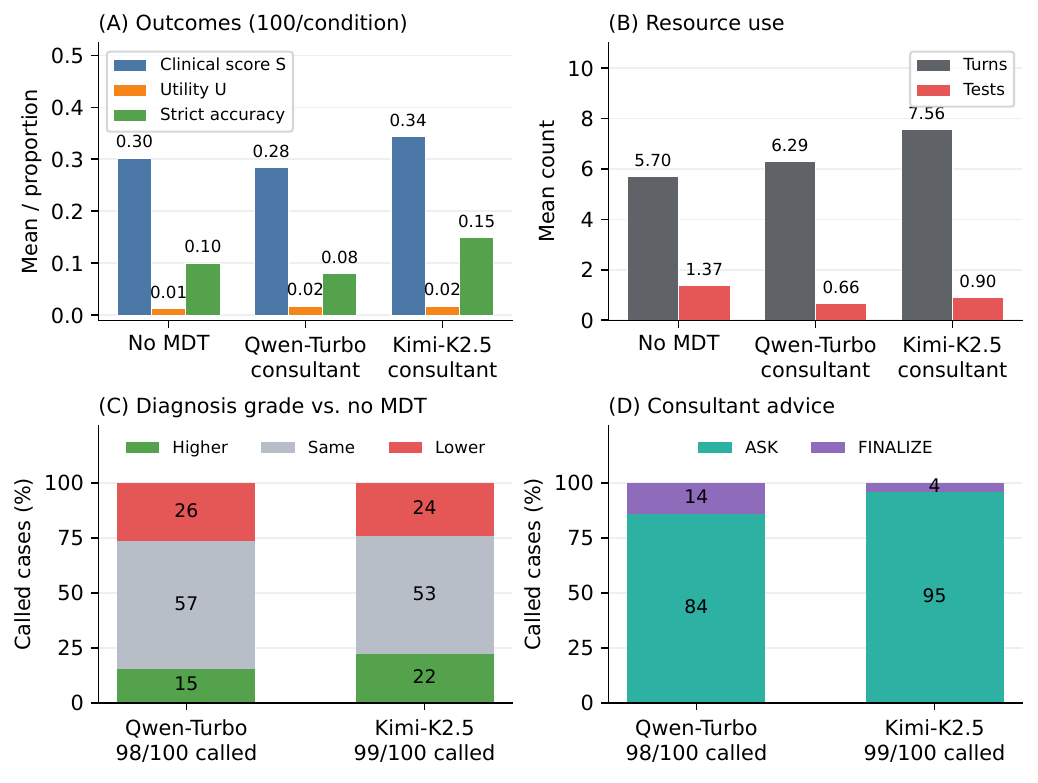}
\caption{MDT decomposition with a Qwen-Turbo primary doctor and Qwen-Turbo or Kimi-K2.5 consultant. (A) Total score, utility, and strict accuracy and (B) counted turns and tests use all 100 assigned episodes per condition. (C) Final diagnosis grade relative to the paired no-MDT episode and (D) consultant advice use the 98 Qwen-consultant and 99 Kimi-consultant episodes that invoked consultation. Bar labels give case counts; percentages use the called-case denominator.}
\label{fig:app_mdt_analysis}
\end{figure}

The Qwen-Turbo MDT study compares no MDT with Qwen-Turbo and Kimi-K2.5 consultants on 100 shared case records with no memory retrieval. Qwen-Turbo is the primary doctor; Kimi-K2.5 provides manager scoring. The Kimi-consultant arm was executed in a separate batch from the no-MDT and Qwen-consultant arms, using the same cases and matched configuration. Table~\ref{tab:app_exp2_full} summarizes clinical scores and examination counts across the three conditions. Figure~\ref{fig:app_mdt_analysis}(A--B) compares outcomes and resource use, while Figure~\ref{fig:app_mdt_analysis}(C--D) shows diagnosis changes and consultant advice among episodes that invoked consultation.

\begin{table}[!htbp]
\centering
\caption{Full MDT summary on 100 shared episodes.}
\label{tab:app_exp2_full}
\small
\begin{tabular}{lccccccc}
\toprule
Setting & Strict & Accept. & $S_d$ & $S_e$ & $S_p$ & $S$ & Tests \\
\midrule
No MDT & 0.100 & 0.180 & 0.283 & 0.271 & 0.400 & 0.303 & 1.37 \\
Qwen-Turbo consultant & 0.080 & 0.130 & 0.230 & 0.301 & 0.395 & 0.284 & \textbf{0.66} \\
Kimi-K2.5 consultant & 0.150 & 0.240 & 0.298 & 0.373 & 0.415 & 0.344 & \textbf{0.90} \\
\bottomrule
\end{tabular}
\end{table}

In this Qwen-Turbo primary-doctor setting, consultation is associated with fewer examination requests and more counted interaction turns. Counted turns exclude the MDT invocation itself. Neither consultant yields a Holm-corrected significant improvement in clinical score, strict accuracy, or utility. All 100 assigned episodes remain in the primary analysis; the Kimi arm contains one recorded doctor error turn.

\paragraph{Stronger primary doctor.}
To examine how MDT affects clinical quality and resource use with a stronger primary doctor, we use GLM-5 as both primary doctor and consultant and compare MDT-enabled and no-MDT conditions on the same 100 cases with memory disabled. Kimi-K2.5 provides manager scoring in both arms. The primary analysis retains all episodes assigned to each condition, including those in which the MDT-enabled doctor does not invoke consultation.
MDT is invoked in 85/100 episodes in the enabled condition.

\begin{table}[!htbp]
\centering
\caption{Strong-primary MDT comparison: GLM-5 primary doctor and GLM-5 consultant, with Kimi-K2.5 manager scoring. All 100 assigned cases per arm enter the analysis; the Holm family contains eleven endpoints.}
\label{tab:app_strong_mdt}
\scriptsize
\setlength{\tabcolsep}{3pt}
\begin{tabular}{lrrrrrr}
\toprule
Endpoint & No MDT & MDT & $\Delta$ & 95\% interval & Raw $p$ & Holm $p$ \\
\midrule
Total score & 0.574 & 0.542 & -0.032 & [-0.083, 0.019] & 0.129 & 1.000 \\
Diagnosis & 0.530 & 0.517 & -0.013 & [-0.087, 0.062] & 0.719 & 1.000 \\
Evidence & 0.564 & 0.493 & -0.070 & [-0.114, -0.028] & 0.006 & 0.056 \\
Plan & 0.700 & 0.675 & -0.025 & [-0.085, 0.033] & 0.476 & 1.000 \\
Utility & 0.165 & 0.139 & -0.026 & [-0.058, 0.005] & 0.135 & 1.000 \\
\rowcolor{TableFocusPurple}
\textbf{Turns} & 4.710 & 6.040 & \textbf{+1.330} & [1.040, 1.630] & $3.10\times10^{-11}$ & $3.41\times10^{-10}$ \\
\rowcolor{TableFocusPurple}
\textbf{Tests} & 2.280 & 1.720 & \textbf{-0.560} & [-0.710, -0.420] & $1.04\times10^{-9}$ & $1.04\times10^{-8}$ \\
Invalid-test rate & 0.090 & 0.078 & -0.012 & [-0.055, 0.035] & 0.436 & 1.000 \\
Strict accuracy & 0.320 & 0.330 & +0.010 & [-0.080, 0.110] & 1.000 & 1.000 \\
Pass rate & 0.450 & 0.410 & -0.040 & [-0.140, 0.060] & 0.541 & 1.000 \\
Non-dangerous diagnosis & 0.800 & 0.780 & -0.020 & [-0.100, 0.060] & 0.804 & 1.000 \\
\bottomrule
\end{tabular}
\end{table}
Table~\ref{tab:app_strong_mdt} shows that MDT increases interaction turns by 1.33 and reduces examinations by 0.56, with both differences passing Holm correction.

\subsection{Longitudinal memory and held-out evaluation}
\label{app:memory_transfer_supp}

\begin{figure}[!htbp]
\centering
\includegraphics[width=\linewidth]{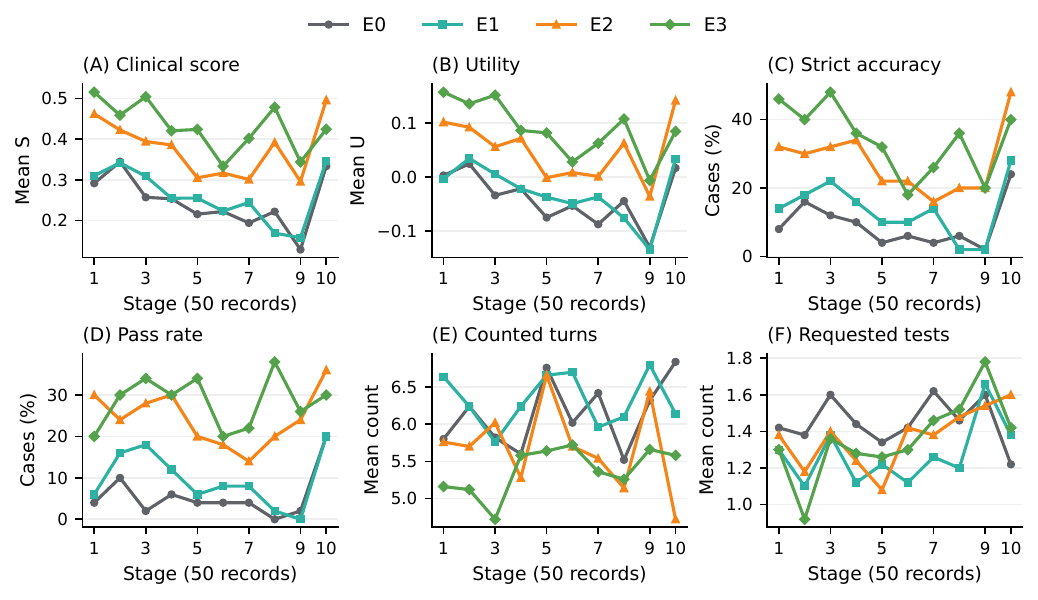}
\caption{Stage-level results across repeated passes of the same 500-record stream. (A) Total score, (B) utility, (C) strict accuracy, (D) pass rate, (E) counted turns, and (F) requested tests are averaged over consecutive windows of 50 records. Within-pass variation also reflects differences between case windows.}
\label{fig:app_learning_curves}
\end{figure}

E0--E3 repeat the same ordered 500-record stream while the memory condition changes. E0 does not retrieve memory; E1 starts with an empty bank and accumulates records online, and E2 and E3 continue from the preceding memory states. This design measures within-stream changes under repeated exposure and accumulating memory. Table~\ref{tab:app_memory_full} summarizes each pass, and Figure~\ref{fig:app_learning_curves} shows stage-level changes over consecutive 50-record windows.

\begin{table}[!htbp]
\centering
\caption{Within-stream memory summary on the same 500 records per pass.}
\label{tab:app_memory_full}
\small
\begin{tabular}{lrrrrrrr}
\toprule
Group & Records & $S$ & $U$ & Strict & Pass & Turns & Tests \\
\midrule
E0 & 500 & 0.246 & -0.040 & 0.092 & 0.056 & 6.134 & 1.450 \\
E1 & 500 & 0.261 & -0.029 & 0.136 & 0.096 & 6.324 & 1.274 \\
E2 & 500 & 0.377 & 0.050 & 0.276 & 0.244 & 5.696 & 1.370 \\
\rowcolor{TableFocusBlue}
\textbf{E3} & 500 & \textbf{0.430} & \textbf{0.089} & \textbf{0.342} & \textbf{0.284} & \textbf{5.380} & 1.360 \\
\bottomrule
\end{tabular}
\end{table}

The held-out experiment pairs H0 and H3 on 200 distinct case contents outside this stream. The doctor, simulator, case order, action space, and scoring rules are matched. H0 runs without memory; H3 starts from the mature stream bank and continues online write-back during the sequence. The following estimates describe the sequential protocol on same-source cases.
\begin{table}[!htbp]
\centering
\caption{Paired H3--H0 results on 200 held-out records. Intervals describe the mean paired difference; raw tests are Wilcoxon for continuous scores and exact McNemar for binary endpoints. Holm correction covers all seven endpoints.}
\label{tab:app_h0h3_paired}
\scriptsize
\setlength{\tabcolsep}{3pt}
\begin{tabular}{lrrrrrr}
\toprule
Endpoint & H0 & H3 & $\Delta$ & 95\% interval & Raw $p$ & Holm $p$ \\
\midrule
Total score & 0.335 & 0.385 & +0.049 & [0.005, 0.094] & 0.120 & 0.370 \\
\rowcolor{TableFocusPurple}
\textbf{Utility} & 0.019 & 0.068 & \textbf{+0.048} & [0.019, 0.078] & 0.003 & \textbf{0.017} \\
\rowcolor{TableFocusPurple}
\textbf{Turns} & 6.645 & 5.730 & \textbf{-0.915} & [-1.290, -0.540] & $3.74\times10^{-6}$ & $2.62\times10^{-5}$ \\
Tests & 1.305 & 1.355 & +0.050 & [-0.060, 0.160] & 0.355 & 0.711 \\
Strict accuracy & 0.260 & 0.295 & +0.035 & [-0.035, 0.105] & 0.392 & 0.711 \\
Pass rate & 0.180 & 0.255 & +0.075 & [0.010, 0.140] & 0.040 & 0.200 \\
Non-dangerous diagnosis & 0.565 & 0.635 & +0.070 & [-0.005, 0.145] & 0.092 & 0.370 \\
\bottomrule
\end{tabular}
\end{table}

\begin{figure}[!htbp]
\centering
\includegraphics[width=\linewidth]{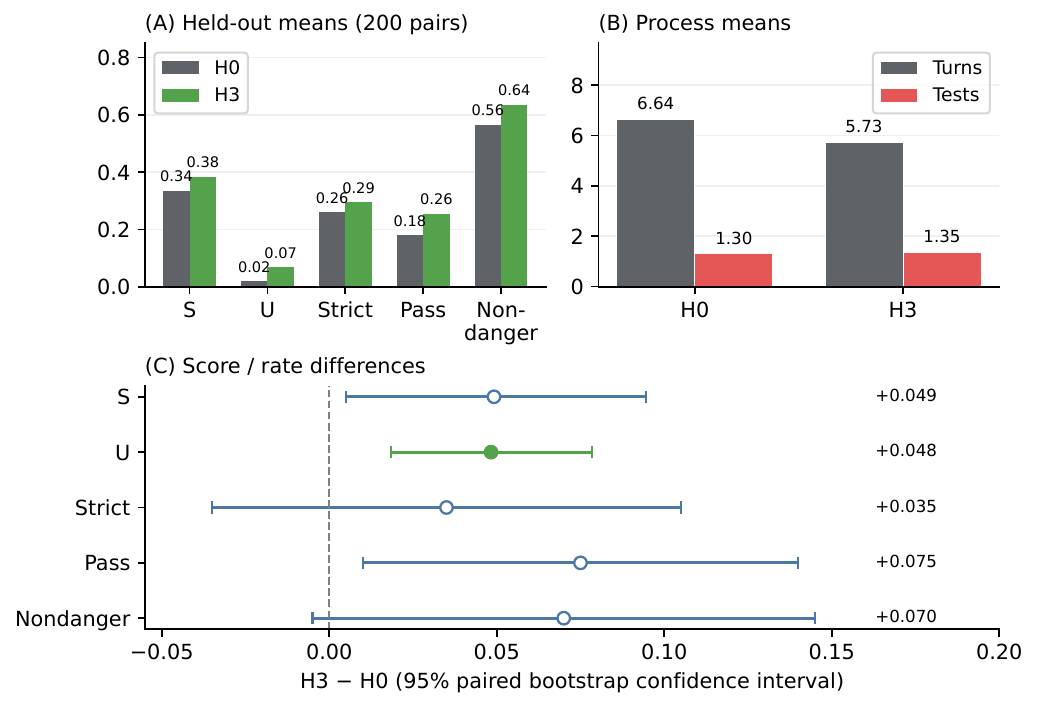}
\caption{Paired H0/H3 results on 200 held-out records. H3 begins with mature memory and allows write-back during evaluation. (A) Outcome and rate means. (B) Turns and tests. (C) Paired mean differences with unadjusted 95\% bootstrap intervals. Solid green markers denote Holm-adjusted $p<0.05$ within the seven-endpoint family; hollow blue markers denote the remaining endpoints.}
\label{fig:app_external_transfer}
\end{figure}

After correction within the seven-endpoint held-out family, H3 uses fewer counted interaction turns and achieves higher composite utility. Full paired estimates are reported in Table~\ref{tab:app_h0h3_paired}; means and difference intervals are shown in Figure~\ref{fig:app_external_transfer}.

\subsection{Update and retention diagnostics}
\label{app:update_retention}

To describe within-pass changes, we compare early and late windows in E0--E3 (Table~\ref{tab:app_early_late}): stages 1--2 form the early window and stages 9--10 the late window, with 50 records per stage and 100 per window.

\begin{figure}[!htbp]
\centering
\includegraphics[width=\linewidth]{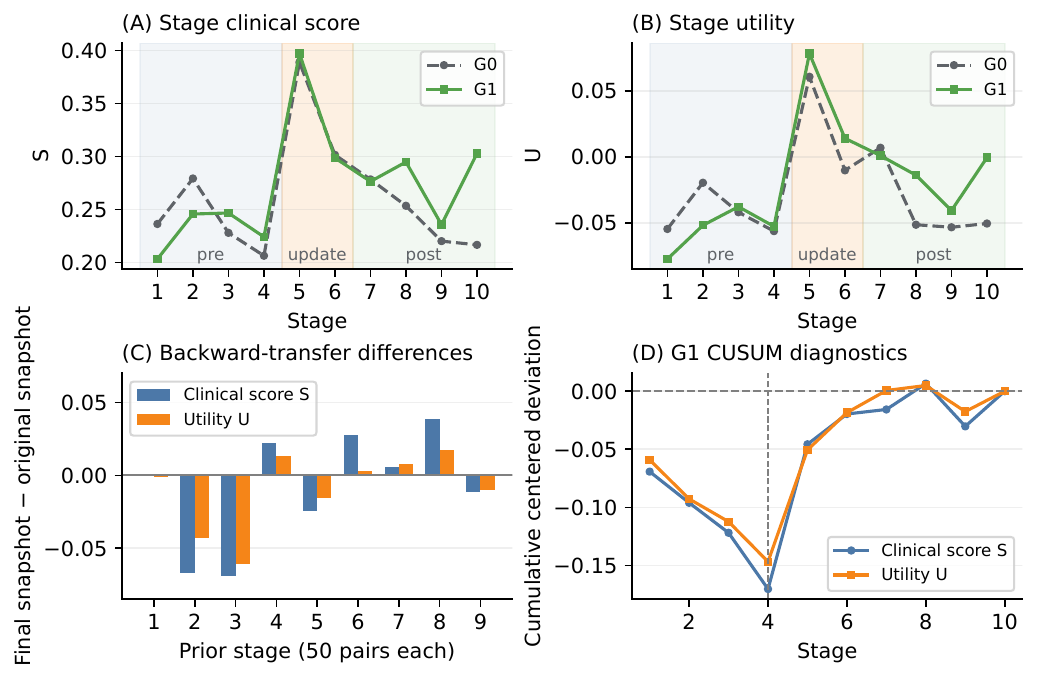}
\caption{Update and retention diagnostics. (A) Stage total score. (B) Stage utility. (C) Score and utility backward-transfer deltas for nine earlier stages. (D) Cumulative centered score and utility deviations in G1 describe variation across stages.}
\label{fig:app_update_diagnostics}
\end{figure}

To examine responses to the update interval and subsequent performance, we compare G0 and G1 before, during, and after the interval. G0 disables memory retrieval, while G1 enables retrieval and online write-back at $K_{\mathrm{ret}}=3$. Both groups follow the same 500-record sequence. Pre-update, update, and post-update windows comprise T1--T4 (200 records), T5--T6 (100 records), and T7--T10 (200 records), respectively. Table~\ref{tab:app_update_bwt} summarizes window means; Figure~\ref{fig:app_update_diagnostics}(A--B) shows stage trajectories, and Figure~\ref{fig:app_update_diagnostics}(D) shows cumulative deviations.

\begin{table}[!htbp]
\centering
\caption{Early--late window means within each pass of the memory stream. Early: stages 1--2; late: stages 9--10; 100 records per window.}
\label{tab:app_early_late}
\small
\begin{tabular}{lcccc}
\toprule
Group & $S$ early & $S$ late & $U$ early & $U$ late \\
\midrule
E0 & 0.318 & 0.231 & 0.013 & -0.057 \\
E1 & 0.326 & 0.252 & 0.015 & -0.050 \\
E2 & 0.442 & 0.396 & 0.097 & 0.053 \\
E3 & 0.487 & 0.384 & 0.146 & 0.039 \\
\bottomrule
\end{tabular}
\end{table}

\begin{table}[!htbp]
\centering
\caption{Update response and backward-transfer diagnostics.}
\label{tab:app_update_bwt}
\small
\begin{tabular}{lcccc}
\toprule
Group/metric & Pre & Update & Post & Change note \\
\midrule
G0 total score & 0.238 & 0.345 & 0.242 & update response $+0.108$ \\
G0 utility & -0.043 & 0.025 & -0.037 & update response $+0.068$ \\
\midrule
G1 total score & 0.230 & 0.348 & 0.277 & update response $+0.118$ \\
G1 utility & -0.055 & 0.046 & -0.013 & update response $+0.101$ \\
\midrule
BWT total score & -- & -- & -- & -0.0087 \\
BWT utility & -- & -- & -- & -0.0099 \\
\bottomrule
\end{tabular}
\end{table}

\paragraph{Descriptive time-series diagnostics.}
Mann--Kendall tests and CUSUM curves are descriptive checks based on ten stage points; the stage-window summaries report pre-update, update, and post-update means. Both G0 and G1 have higher means in the update window, characterizing their response to the shifted case distribution.

The update interval produces a positive current-stage response in this probe. To assess how later updates affect earlier-stage performance, we compare the final and original-stage snapshots on 450 paired records. BWT has a small negative point estimate; estimates and intervals are reported in Table~\ref{tab:app_bwt_inference}, and differences for individual earlier stages are shown in Figure~\ref{fig:app_update_diagnostics}(C).
\begin{table}[!htbp]
\centering
\caption{Backward-transfer contrasts on 450 paired records: final snapshot minus the original-stage snapshot for each record.}
\label{tab:app_bwt_inference}
\scriptsize
\setlength{\tabcolsep}{3pt}
\begin{tabular}{lrrrr}
\toprule
Endpoint & Mean difference & 95\% interval & Raw $p$ & Holm $p$ \\
\midrule
Total score & -0.0087 & [-0.034, 0.017] & 0.920 & 0.920 \\
Utility & -0.0099 & [-0.027, 0.008] & 0.368 & 0.736 \\
\bottomrule
\end{tabular}
\end{table}

\subsection{Memory-content and retrieval-scale ablations}
\label{app:curves_ablations}

\begin{figure}[!htbp]
\centering
\includegraphics[width=\linewidth]{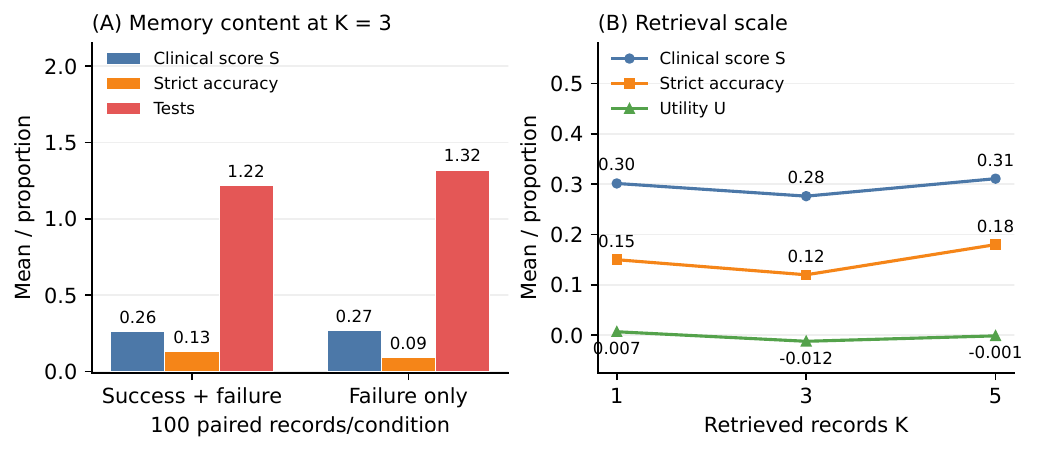}
\caption{Memory ablations on the 100-record mechanism subset. (A) Total score, strict accuracy, and requested tests for success-plus-failure versus failure-only memory at $K_{\mathrm{ret}}=3$. (B) Total score, utility, and strict accuracy across $K_{\mathrm{ret}}\in\{1,3,5\}$.}
\label{fig:app_ablation_results}
\end{figure}

To examine how memory composition and the number of retrieved cards affect performance, we conduct memory-content and retrieval-scale ablations on the same ordered 100-case subset. The former compares success-plus-failure with failure-only memory at $K_{\mathrm{ret}}=3$; the latter varies $K_{\mathrm{ret}}$ with success and failure memories enabled. The two ablations were executed separately with independent memory banks, so each has its own success-plus-failure $K_{\mathrm{ret}}=3$ run. Memory-content comparisons appear in Table~\ref{tab:app_ablation_memory_type} and Figure~\ref{fig:app_ablation_results}(A); retrieval-scale comparisons appear in Table~\ref{tab:app_ablation_topk} and Figure~\ref{fig:app_ablation_results}(B).

\begin{table}[!htbp]
\centering
\caption{Memory-content ablation on the 100-case mechanism subset with $K_{\mathrm{ret}}=3$.}
\label{tab:app_ablation_memory_type}
\small
\begin{tabular}{lcccccc}
\toprule
Memory setting & $S$ & $U$ & Pass & Turns & Tests & Strict \\
\midrule
Success + failure & 0.2627 & -0.0217 & 0.09 & 6.21 & 1.22 & 0.13 \\
Failure only & 0.2668 & -0.0179 & 0.09 & 5.99 & 1.32 & 0.09 \\
\bottomrule
\end{tabular}
\end{table}

\begin{table}[!htbp]
\centering
\caption{Retrieval-scale ablation on the 100-case mechanism subset with success and failure memories enabled.}
\label{tab:app_ablation_topk}
\small
\begin{tabular}{ccccccc}
\toprule
$K_{\mathrm{ret}}$ & $S$ & $U$ & Pass & Turns & Tests & Strict \\
\midrule
1 & 0.3015 & 0.0068 & 0.13 & 5.68 & 1.38 & 0.15 \\
3 & 0.2763 & -0.0122 & 0.08 & 6.28 & 1.21 & 0.12 \\
5 & 0.3110 & -0.0013 & 0.17 & 6.39 & 1.34 & 0.18 \\
\bottomrule
\end{tabular}
\end{table}

The point estimates vary across memory-content and retrieval-depth settings. After the corresponding family corrections, the retrieval-depth analysis does not detect a reliable endpoint difference at this scale.
\paragraph{Frozen-bank retrieval control.}\label{app:memory_mechanism}
To examine how the relevance of retrieved content to the current case affects performance, we compare relevant and case-mismatched retrieval from the same frozen memory bank on the same 100 case records, holding card count and truncation policy fixed and including a no-memory reference. Table~\ref{tab:app_frozen_memory} reports the relevant-versus-mismatched comparison; complete comparisons with the no-memory reference appear in \href{\CodeURL/blob/main/results/final_holm_tests.csv}{\nolinkurl{results/final_holm_tests.csv}} in the code package.

\begin{table}[!htbp]
\centering
\caption{Relevant minus case-mismatched retrieval in the frozen-bank control. Each condition has 100 records, and Holm correction covers nine endpoints.}
\label{tab:app_frozen_memory}
\scriptsize
\setlength{\tabcolsep}{3pt}
\begin{tabular}{lrrrrrr}
\toprule
Endpoint & Mismatched & Relevant & $\Delta$ & 95\% interval & Raw $p$ & Holm $p$ \\
\midrule
Total score & 0.330 & 0.393 & +0.062 & [-0.006, 0.131] & 0.099 & 0.520 \\
Utility & 0.018 & 0.063 & +0.045 & [-0.001, 0.092] & 0.087 & 0.520 \\
Turns & 6.640 & 5.940 & -0.700 & [-1.240, -0.160] & 0.020 & 0.159 \\
Tests & 1.220 & 1.500 & +0.280 & [0.130, 0.430] & $5.97\times10^{-4}$ & 0.005 \\
Evidence & 0.275 & 0.298 & +0.023 & [-0.018, 0.065] & 0.229 & 0.702 \\
Plan & 0.290 & 0.380 & +0.090 & [0.015, 0.165] & 0.021 & 0.159 \\
Strict accuracy & 0.250 & 0.300 & +0.050 & [-0.050, 0.150] & 0.424 & 0.702 \\
Pass rate & 0.180 & 0.250 & +0.070 & [-0.030, 0.170] & 0.248 & 0.702 \\
Non-dangerous diagnosis & 0.560 & 0.650 & +0.090 & [-0.020, 0.200] & 0.175 & 0.702 \\
\bottomrule
\end{tabular}
\end{table}
Relative to mismatched retrieval, relevant retrieval increases mean total score by 0.062 and utility by 0.045, and reduces interaction turns by 0.700. After Holm correction across nine endpoints, the increase of 0.280 in examination count is the only significant difference.

\section{Physician audits and scorer robustness}
\label{app:llm_sim_audit}

\FloatBarrier
\subsection{Two-physician output-scoring audit}
To assess agreement between automatic scoring and physician judgments, two physicians independently reviewed the same 80 episode outputs, without access to one another's ratings or the automatic scoring key. They assessed diagnosis, supporting evidence, and management against the supplied reference material. The output sample spans model-comparison, stream-memory, and held-out conditions. The sample size is 80 outputs, each rated by two physicians.

For the reported diagnosis agreement, \emph{exact} and \emph{mostly correct} grades form the positive class; partially correct, wrong but low-risk, and dangerous grades form the negative class. Cohen's $\kappa$ is calculated on these binary labels (Table~\ref{tab:app_llm_sim_audit}). The accompanying audit records retain the individual grades, evidence scores, and plan scores.

\begin{table}[!htbp]
\centering
\caption{Binary diagnosis agreement on the same 80 outputs. The positive class is exact or mostly correct.}
\label{tab:app_llm_sim_audit}
\small
\begin{tabular}{lc}
\toprule
Comparison & Cohen's $\kappa$ \\
\midrule
Physician A vs.\ physician B & 0.824 \\
Physician A vs.\ automatic scorer & 0.727 \\
Physician B vs.\ automatic scorer & 0.802 \\
\bottomrule
\end{tabular}
\end{table}

The 80 outputs comprise E0 (10), E3 (22), H0 (10), H3 (14), Qwen-Turbo (6), Qwen3.5 (4), GLM-5 (7), and Kimi-K2.5 (7). These are descriptive audit strata with unequal counts. Binary agreement is 91.25\% between physicians and 86.25\%/90.00\% between each physician and the automatic key. Exact agreement on all five diagnosis grades is 71.25\% between physicians.

\begin{figure}[!htbp]
\centering
\includegraphics[width=\linewidth]{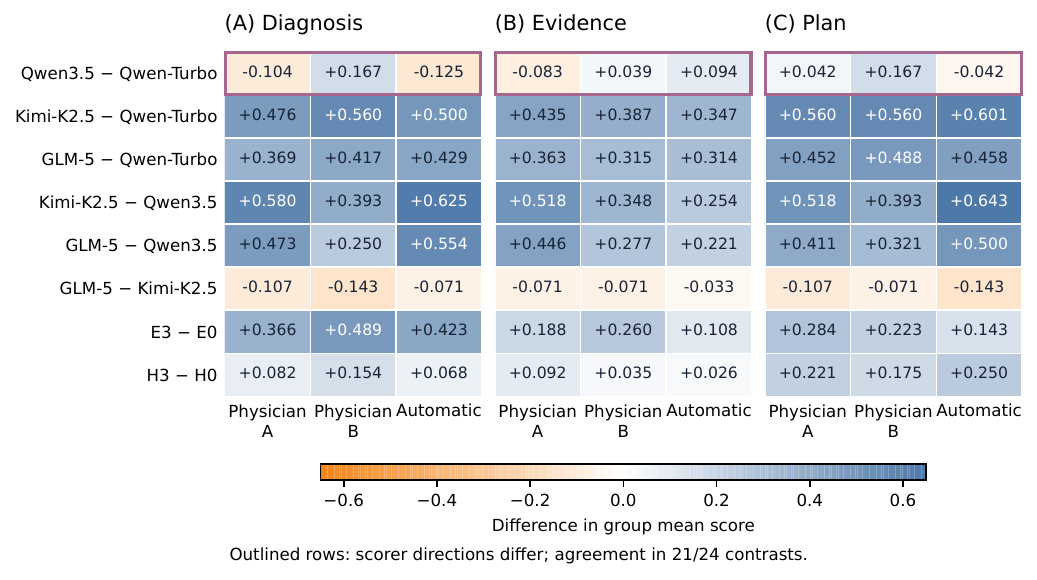}
\caption{All 24 physician-audit direction comparisons across diagnosis, evidence, and management. Each cell gives a signed difference in sample means for physician A, physician B, or the automatic scorer. Outlined rows mark the three comparisons with discordant directions; the remaining 21 agree across all three scorers. Qwen3.5 abbreviates Qwen3.5-122B-A10B.}
\label{fig:app_audit_directions}
\end{figure}

The 24 direction comparisons consist of all six model pairs plus E3--E0 and H3--H0, each evaluated on diagnosis, evidence, and plan means. A zero difference is treated as its own direction after rounding at 12 decimal places. All three scorers agree on the sign in 21/24 comparisons, including all three dimensions for E3--E0 and H3--H0. Direction agreement is calculated from the audit-group means; all comparisons are shown in Figure~\ref{fig:app_audit_directions}.

\subsection{Source-to-episode transformation audit}
\label{app:transformation_audit}
\label{app:conversion_audit}
To assess preservation of the source medical information during case conversion, two physicians reviewed the same 50 source-to-episode conversions. Each compared the MedQA source with the generated patient, examination, and manager views, considering target-diagnosis preservation, separation of history and examination information, test-catalog plausibility, critical-information fabrication, and overall faithfulness. This audit concerns case construction and is distinct from the 80-output scoring audit.

\begin{table}[!htbp]
\centering
\caption{Physician assessment of 50 MedQA-to-episode conversions. Overall faithfulness uses a 1--5 scale.}
\label{tab:app_transformation_audit}
\small
\begin{tabular}{lcc}
\toprule
Criterion & Physician A & Physician B \\
\midrule
Target diagnosis preserved & 50/50 & 50/50 \\
History/examination separation faithful & 50/50 & 45/50 \\
Test catalog clinically plausible & 50/50 & 50/50 \\
No fabricated critical information & 48/50 & 50/50 \\
Mean overall faithfulness & 4.78 & 4.46 \\
\bottomrule
\end{tabular}
\end{table}

Table~\ref{tab:app_transformation_audit} shows that both physicians confirmed target-diagnosis preservation in all 50 cases, with mean overall faithfulness of 4.78 and 4.46, respectively. Physician A identified possible critical-information fabrication in two cases.

\subsection{Alternative judges and scoring-path checks}
\label{app:judge_robustness}
To assess the dependence of the main results on the manager judge, we re-score saved final outputs with DeepSeek-V3.2 and Qwen-Max and repeat scoring with the original judge. We also compute a deterministic diagnosis-match rate: the fraction of normalized diagnoses that match the acceptable reference set. The main scoring procedure first matches the diagnosis and then lowers its grade when the management plan triggers the major-strategy-deviation rule; the diagnosis-matching check retains the first step. This comparison measures how the management-dependent grading step affects the reported strict accuracy.
The two replacement-judge analyses each cover 2,520 saved outputs: 320 four-model outputs, 2,000 E0--E3 stream outputs, and 200 outputs from a held-out subset. The same-judge retest covers 520 outputs (the model-comparison outputs and that held-out subset). The held-out scoring subset comprises 100 H0/H3 case pairs.

Scoring agreement and model-ranking results are reported in Table~\ref{tab:app_judge_agreement}.

\begin{table}[!htbp]
\centering
\caption{Scoring agreement with the original automatic scorer. Correlations concern total score; $\kappa$ concerns strict diagnostic correctness.}
\label{tab:app_judge_agreement}
\scriptsize
\setlength{\tabcolsep}{3pt}
\begin{tabular}{lrrrrr}
\toprule
Judge/check & Outputs & Score $r$ & Strict $\kappa$ & Rank $\tau$ & Extraction fallback \\
\midrule
DeepSeek-V3.2 & 2520 & 0.901 & 0.868 & \textbf{1.0} & 0 \\
Qwen-Max & 2520 & 0.878 & 0.875 & \textbf{1.0} & 107 \\
Kimi-K2.5 retest & 520 & 0.962 & 0.965 & \textbf{1.0} & 0 \\
\bottomrule
\end{tabular}
\end{table}
Both replacement pipelines preserve the four-model total-score ordering GLM-5 $>$ Kimi-K2.5 $>$ Qwen3.5 $>$ Qwen-Turbo and the increasing E0--E3 mean-score pattern in their covered outputs. For 107/2,520 Qwen-Max outputs, failed LLM extraction triggers deterministic substring matching of rubric points in the evidence and plan/follow-up text. These outputs remain in the reported results, so the Qwen-Max comparison measures the complete scoring pipeline, including this fallback.

On the 100 paired cases in the rescored held-out subset, both replacement judges preserve a positive mean utility difference for H3 relative to H0 (Table~\ref{tab:app_judge_heldout_utility}). Utility is recomputed from each judge's outcome score with the recorded turns, tests, and invalid-test rate held fixed.

\begin{table}[!htbp]
\centering
\caption{Mean utility on the 100-pair held-out scoring-robustness subset. Differences are descriptive; the final 200-pair primary analysis is reported in Appendix~\ref{app:memory_transfer_supp}.}
\label{tab:app_judge_heldout_utility}
\scriptsize
\setlength{\tabcolsep}{5pt}
\begin{tabular}{lrrrr}
\toprule
Judge/check & Pairs & H0 $U$ & H3 $U$ & H3$-$H0 \\
\midrule
DeepSeek-V3.2 & 100 & -0.0002 & 0.0744 & \textbf{+0.0746} \\
Qwen-Max & 100 & 0.0383 & 0.1025 & \textbf{+0.0641} \\
Kimi-K2.5 retest & 100 & 0.0300 & 0.0768 & \textbf{+0.0468} \\
\bottomrule
\end{tabular}
\end{table}

\begin{figure}[!htbp]
\centering
\includegraphics[width=\linewidth]{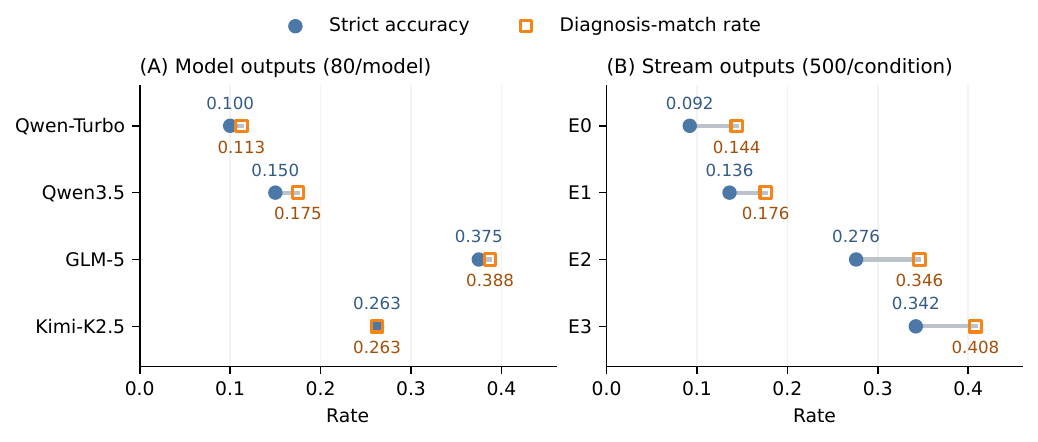}
\caption{Strict accuracy and diagnosis-match rate on the same model and stream outputs (80 outputs per model; 500 per stream condition). Diagnosis matching uses the normalized diagnosis; strict accuracy additionally applies the management-dependent grading step. Points and labels show each rate, and connecting segments show the difference between the two definitions.}
\label{fig:app_judge_free}
\end{figure}

The management-dependent grading step changes the strict-correctness label in 127/2,520 outputs (5.04\%). The diagnosis-match rates preserve the four-model ordering and the increasing E0--E3 pattern (Figure~\ref{fig:app_judge_free}). Case-level results for all covered conditions are included in the accompanying results.

These checks assess ranking and directional stability on the covered outputs.

\subsection{Complete corrected statistical comparisons}
\label{app:complete_tests}
Table~\ref{tab:app_holm_families} identifies each correction family. All 141 tests are provided in \href{\CodeURL/blob/main/results/final_holm_tests.csv}{\nolinkurl{results/final_holm_tests.csv}} in the code package. For paired comparisons, the effect is the first named condition minus the second. Continuous endpoints use paired Wilcoxon tests, and binary Strict, Pass, Non-danger, and Any error use exact McNemar tests. Retrieval-scale rows use Friedman (continuous) or Cochran $Q$ (binary) omnibus tests and therefore have no single paired effect or interval. G0/G1 stage-trend effects are Kendall $\tau$, with no record-level bootstrap interval. They summarize only ten stage observations and remain descriptive.

Abbreviations are QT: Qwen-Turbo; Q3: Qwen3.5-122B-A10B; G: GLM-5; K: Kimi-K2.5; A/B/C: no MDT/Qwen-Turbo consultant/Kimi-K2.5 consultant in the Qwen-Turbo study (A/B/C are the retained statistical condition codes); Rel/Mis/None: relevant/mismatched/no memory. Full--failure-only denotes the memory-content ablation. Norm--Raw denotes the interface intervention within the named model. Intervals are unadjusted 95\% paired percentile-bootstrap intervals; significance is determined by the Holm-adjusted tests.

\begin{table}[!htbp]
\centering
\caption{Holm families in the final statistical analysis. Counts refer to endpoints and contrasts within each scientific question; adjustment is within families. Sig. counts Holm-adjusted $p<0.05$.}
\label{tab:app_holm_families}
\scriptsize
\setlength{\tabcolsep}{3pt}
\begin{tabular}{llrrllrr}
\toprule
ID & Question & Tests & Sig. & ID & Question & Tests & Sig. \\
\midrule
F1 & Held-out H3 vs. H0 & 7 & 2 & F10 & Retrieval $K=1,3,5$ & 6 & 0 \\
F2 & Models vs. Qwen-Turbo & 9 & 6 & F11 & G0 trends (descriptive) & 6 & 0 \\
F3 & MDT arms vs. no MDT & 10 & 3 & F12 & G1 trends (descriptive) & 6 & 0 \\
F4 & Kimi vs. Qwen consultant & 5 & 2 & F13 & Final vs. original (450 pairs) & 2 & 0 \\
F5 & Relevant vs. mismatched memory & 9 & 1 & F14 & Norm.--raw: outcomes & 12 & 2 \\
F6 & Relevant vs. no memory & 9 & 0 & F15 & Norm.--raw: process & 8 & 2 \\
F7 & Mismatched vs. no memory & 9 & 0 & F16 & Norm.--raw: interface errors & 8 & 2 \\
F8 & Strong-primary MDT (all cases) & 11 & 2 & F17 & Raw models vs. Qwen-Turbo & 9 & 5 \\
F9 & Full vs. failure-only memory & 6 & 0 & F18 & Normalized models vs. Qwen-Turbo & 9 & 8 \\
\bottomrule
\end{tabular}
\end{table}
\FloatBarrier

\clearpage
\section*{NeurIPS Paper Checklist}

\begin{enumerate}

\Needspace{4\baselineskip}
\item {\bf Claims}
    \item[] Question: Do the main claims made in the abstract and introduction accurately reflect the paper's contributions and scope?
    \item[] Answer: \answerYes{}
    \item[] Justification: The abstract and Introduction describe an executable, simulation-based evaluation protocol. The empirical claims distinguish repeated exposure within a fixed stream from sequential evaluation on 200 held-out records, and limit the positive held-out finding to turns and composite utility after multiplicity correction.
    \Needspace{3\baselineskip}
    \item[] Guidelines:
    \begin{itemize}
        \item The answer \answerNA{} means that the abstract and introduction do not include the claims made in the paper.
        \item The abstract and/or introduction should clearly state the claims made, including the contributions made in the paper and important assumptions and limitations. A \answerNo{} or \answerNA{} answer to this question will not be perceived well by the reviewers. 
        \item The claims made should match theoretical and experimental results, and reflect how much the results can be expected to generalize to other settings. 
        \item It is fine to include aspirational goals as motivation as long as it is clear that these goals are not attained by the paper. 
    \end{itemize}

\Needspace{4\baselineskip}
\item {\bf Limitations}
    \item[] Question: Does the paper discuss the limitations of the work performed by the authors?
    \item[] Answer: \answerYes{}
    \item[] Justification: The paper, appendices, and data card discuss simulation and single-source scope, content duplicates, sequential memory write-back, scorer uncertainty, metric-weight choices, non-significant clinical endpoints, and the limits of the update and retrieval controls.
    \Needspace{3\baselineskip}
    \item[] Guidelines:
    \begin{itemize}
        \item The answer \answerNA{} means that the paper has no limitation while the answer \answerNo{} means that the paper has limitations, but those are not discussed in the paper. 
        \item The authors are encouraged to create a separate ``Limitations'' section in their paper.
        \item The paper should point out any strong assumptions and how robust the results are to violations of these assumptions (e.g., independence assumptions, noiseless settings, model well-specification, asymptotic approximations only holding locally). The authors should reflect on how these assumptions might be violated in practice and what the implications would be.
        \item The authors should reflect on the scope of the claims made, e.g., if the approach was only tested on a few datasets or with a few runs. In general, empirical results often depend on implicit assumptions, which should be articulated.
        \item The authors should reflect on the factors that influence the performance of the approach. For example, a facial recognition algorithm may perform poorly when image resolution is low or images are taken in low lighting. Or a speech-to-text system might not be used reliably to provide closed captions for online lectures because it fails to handle technical jargon.
        \item The authors should discuss the computational efficiency of the proposed algorithms and how they scale with dataset size.
        \item If applicable, the authors should discuss possible limitations of their approach to address problems of privacy and fairness.
        \item While the authors might fear that complete honesty about limitations might be used by reviewers as grounds for rejection, a worse outcome might be that reviewers discover limitations that aren't acknowledged in the paper. The authors should use their best judgment and recognize that individual actions in favor of transparency play an important role in developing norms that preserve the integrity of the community. Reviewers will be specifically instructed to not penalize honesty concerning limitations.
    \end{itemize}

\Needspace{4\baselineskip}
\item {\bf Theory assumptions and proofs}
    \item[] Question: For each theoretical result, does the paper provide the full set of assumptions and a complete (and correct) proof?
    \item[] Answer: \answerNA{}
    \item[] Justification: The paper does not present theoretical theorems or proofs. The mathematical definitions specify evaluation metrics and statistical quantities used by the benchmark, with their assumptions and interpretations described in the evaluation protocol.
    \Needspace{3\baselineskip}
    \item[] Guidelines:
    \begin{itemize}
        \item The answer \answerNA{} means that the paper does not include theoretical results. 
        \item All the theorems, formulas, and proofs in the paper should be numbered and cross-referenced.
        \item All assumptions should be clearly stated or referenced in the statement of any theorems.
        \item The proofs can either appear in the main paper or the supplemental material, but if they appear in the supplemental material, the authors are encouraged to provide a short proof sketch to provide intuition. 
        \item Inversely, any informal proof provided in the core of the paper should be complemented by formal proofs provided in appendix or supplemental material.
        \item Theorems and Lemmas that the proof relies upon should be properly referenced. 
    \end{itemize}

    \item {\bf Experimental result reproducibility}
    \item[] Question: Does the paper fully disclose all the information needed to reproduce the main experimental results of the paper to the extent that it affects the main claims and/or conclusions of the paper (regardless of whether the code and data are provided or not)?
    \item[] Answer: \answerYes{}
    \item[] Justification: The paper specifies role visibility, action and error handling, resource limits, memory retrieval/write-back, metric definitions, and statistical families. The accompanying package supplies frozen records and offline analysis scripts. Repeating model inference additionally requires access to the named external APIs and may be affected by provider-side changes.
    \Needspace{3\baselineskip}
    \item[] Guidelines:
    \begin{itemize}
        \item The answer \answerNA{} means that the paper does not include experiments.
        \item If the paper includes experiments, a \answerNo{} answer to this question will not be perceived well by the reviewers: Making the paper reproducible is important, regardless of whether the code and data are provided or not.
        \item If the contribution is a dataset and\slash or model, the authors should describe the steps taken to make their results reproducible or verifiable. 
        \item Depending on the contribution, reproducibility can be accomplished in various ways. For example, if the contribution is a novel architecture, describing the architecture fully might suffice, or if the contribution is a specific model and empirical evaluation, it may be necessary to either make it possible for others to replicate the model with the same dataset, or provide access to the model. In general. releasing code and data is often one good way to accomplish this, but reproducibility can also be provided via detailed instructions for how to replicate the results, access to a hosted model (e.g., in the case of a large language model), releasing of a model checkpoint, or other means that are appropriate to the research performed.
        \item While NeurIPS does not require releasing code, the conference does require all submissions to provide some reasonable avenue for reproducibility, which may depend on the nature of the contribution. For example
        \begin{enumerate}
            \item If the contribution is primarily a new algorithm, the paper should make it clear how to reproduce that algorithm.
            \item If the contribution is primarily a new model architecture, the paper should describe the architecture clearly and fully.
            \item If the contribution is a new model (e.g., a large language model), then there should either be a way to access this model for reproducing the results or a way to reproduce the model (e.g., with an open-source dataset or instructions for how to construct the dataset).
            \item We recognize that reproducibility may be tricky in some cases, in which case authors are welcome to describe the particular way they provide for reproducibility. In the case of closed-source models, it may be that access to the model is limited in some way (e.g., to registered users), but it should be possible for other researchers to have some path to reproducing or verifying the results.
        \end{enumerate}
    \end{itemize}

\Needspace{4\baselineskip}
\item {\bf Open access to data and code}
    \item[] Question: Does the paper provide open access to the data and code, with sufficient instructions to faithfully reproduce the main experimental results, as described in supplemental material?
    \item[] Answer: \answerYes{}
    \item[] Justification: The public code repository and dataset are linked in Section~\ref{sec:framework}. They provide the episode engine, configurations, paper analysis inputs, data records, schemas, license notices, and instructions for offline reproduction of the reported tables and figures.
    \Needspace{3\baselineskip}
    \item[] Guidelines:
    \begin{itemize}
        \item The answer \answerNA{} means that paper does not include experiments requiring code.
        \item Please see the NeurIPS code and data submission guidelines (\url{https://neurips.cc/public/guides/CodeSubmissionPolicy}) for more details.
        \item While we encourage the release of code and data, we understand that this might not be possible, so \answerNo{} is an acceptable answer. Papers cannot be rejected simply for not including code, unless this is central to the contribution (e.g., for a new open-source benchmark).
        \item The instructions should contain the exact command and environment needed to run to reproduce the results. See the NeurIPS code and data submission guidelines (\url{https://neurips.cc/public/guides/CodeSubmissionPolicy}) for more details.
        \item The authors should provide instructions on data access and preparation, including how to access the raw data, preprocessed data, intermediate data, and generated data, etc.
        \item The authors should provide scripts to reproduce all experimental results for the new proposed method and baselines. If only a subset of experiments are reproducible, they should state which ones are omitted from the script and why.
        \item At submission time, to preserve anonymity, the authors should release anonymized versions (if applicable).
        \item Providing as much information as possible in supplemental material (appended to the paper) is recommended, but including URLs to data and code is permitted.
    \end{itemize}

\Needspace{4\baselineskip}
\item {\bf Experimental setting/details}
    \item[] Question: Does the paper specify all the training and test details (e.g., data splits, hyperparameters, how they were chosen, type of optimizer) necessary to understand the results?
    \item[] Answer: \answerYes{}
    \item[] Justification: The experiment overview and Appendix~\ref{app:splits_repro} state sample sizes, model roles, pairing, action budgets, memory states, and retrieval/write-back behavior. The statistical manifest identifies the records and comparisons used for each analysis.
    \Needspace{3\baselineskip}
    \item[] Guidelines:
    \begin{itemize}
        \item The answer \answerNA{} means that the paper does not include experiments.
        \item The experimental setting should be presented in the core of the paper to a level of detail that is necessary to appreciate the results and make sense of them.
        \item The full details can be provided either with the code, in appendix, or as supplemental material.
    \end{itemize}

\Needspace{4\baselineskip}
\item {\bf Experiment statistical significance}
    \item[] Question: Does the paper report error bars suitably and correctly defined or other appropriate information about the statistical significance of the experiments?
    \item[] Answer: \answerYes{}
    \item[] Justification: Appendix~\ref{app:statistical_tests} defines paired Wilcoxon and exact McNemar tests, paired bootstrap intervals, and within-family Holm correction. Final tables report effect estimates with raw and adjusted $p$-values. The text distinguishes record-level uncertainty, exploratory analyses, and results that do not establish an effect or equivalence.
    \Needspace{3\baselineskip}
    \item[] Guidelines:
    \begin{itemize}
        \item The answer \answerNA{} means that the paper does not include experiments.
        \item The authors should answer \answerYes{} if the results are accompanied by error bars, confidence intervals, or statistical significance tests, at least for the experiments that support the main claims of the paper.
        \item The factors of variability that the error bars are capturing should be clearly stated (for example, train/test split, initialization, random drawing of some parameter, or overall run with given experimental conditions).
        \item The method for calculating the error bars should be explained (closed form formula, call to a library function, bootstrap, etc.)
        \item The assumptions made should be given (e.g., Normally distributed errors).
        \item It should be clear whether the error bar is the standard deviation or the standard error of the mean.
        \item It is OK to report 1-sigma error bars, but one should state it. The authors should preferably report a 2-sigma error bar than state that they have a 96\% CI, if the hypothesis of Normality of errors is not verified.
        \item For asymmetric distributions, the authors should be careful not to show in tables or figures symmetric error bars that would yield results that are out of range (e.g., negative error rates).
        \item If error bars are reported in tables or plots, the authors should explain in the text how they were calculated and reference the corresponding figures or tables in the text.
    \end{itemize}

\Needspace{4\baselineskip}
\item {\bf Experiments compute resources}
    \item[] Question: For each experiment, does the paper provide sufficient information on the computer resources (type of compute workers, memory, time of execution) needed to reproduce the experiments?
    \item[] Answer: \answerNo{}
    \item[] Justification: The paper reports API-based inference and the accompanying resource ledger summarizes the available calls, tokens, and timing records by experiment. Complete per-experiment execution-time and compute-resource records are unavailable.
    \Needspace{3\baselineskip}
    \item[] Guidelines:
    \begin{itemize}
        \item The answer \answerNA{} means that the paper does not include experiments.
        \item The paper should indicate the type of compute workers CPU or GPU, internal cluster, or cloud provider, including relevant memory and storage.
        \item The paper should provide the amount of compute required for each of the individual experimental runs as well as estimate the total compute. 
        \item The paper should disclose whether the full research project required more compute than the experiments reported in the paper (e.g., preliminary or failed experiments that didn't make it into the paper). 
    \end{itemize}
    
\Needspace{4\baselineskip}
\item {\bf Code of ethics}
    \item[] Question: Does the research conducted in the paper conform, in every respect, with the NeurIPS Code of Ethics \url{https://neurips.cc/public/EthicsGuidelines}?
    \item[] Answer: \answerYes{}
    \item[] Justification: The work evaluates simulated benchmark cases and does not claim clinical deployment readiness. The paper and artifact document medical over-interpretation risks, source provenance, licensing boundaries, expert-audit procedures, and limits on the use of the released material.
    \Needspace{3\baselineskip}
    \item[] Guidelines:
    \begin{itemize}
        \item The answer \answerNA{} means that the authors have not reviewed the NeurIPS Code of Ethics.
        \item If the authors answer \answerNo, they should explain the special circumstances that require a deviation from the Code of Ethics.
        \item The authors should make sure to preserve anonymity (e.g., if there is a special consideration due to laws or regulations in their jurisdiction).
    \end{itemize}

\Needspace{4\baselineskip}
\item {\bf Broader impacts}
    \item[] Question: Does the paper discuss both potential positive societal impacts and negative societal impacts of the work performed?
    \item[] Answer: \answerYes{}
    \item[] Justification: The discussion and limitations section covers positive impacts, such as more transparent and process-aware medical-agent evaluation, and negative risks, such as over-interpreting simulated benchmark results as evidence of clinical readiness. The paper explicitly states that the benchmark alone should not be used to justify autonomous clinical deployment.
    \Needspace{3\baselineskip}
    \item[] Guidelines:
    \begin{itemize}
        \item The answer \answerNA{} means that there is no societal impact of the work performed.
        \item If the authors answer \answerNA{} or \answerNo, they should explain why their work has no societal impact or why the paper does not address societal impact.
        \item Examples of negative societal impacts include potential malicious or unintended uses (e.g., disinformation, generating fake profiles, surveillance), fairness considerations (e.g., deployment of technologies that could make decisions that unfairly impact specific groups), privacy considerations, and security considerations.
        \item The conference expects that many papers will be foundational research and not tied to particular applications, let alone deployments. However, if there is a direct path to any negative applications, the authors should point it out. For example, it is legitimate to point out that an improvement in the quality of generative models could be used to generate Deepfakes for disinformation. On the other hand, it is not needed to point out that a generic algorithm for optimizing neural networks could enable people to train models that generate Deepfakes faster.
        \item The authors should consider possible harms that could arise when the technology is being used as intended and functioning correctly, harms that could arise when the technology is being used as intended but gives incorrect results, and harms following from (intentional or unintentional) misuse of the technology.
        \item If there are negative societal impacts, the authors could also discuss possible mitigation strategies (e.g., gated release of models, providing defenses in addition to attacks, mechanisms for monitoring misuse, mechanisms to monitor how a system learns from feedback over time, improving the efficiency and accessibility of ML).
    \end{itemize}
    
\Needspace{4\baselineskip}
\item {\bf Safeguards}
    \item[] Question: Does the paper describe safeguards that have been put in place for responsible release of data or models that have a high risk for misuse (e.g., pre-trained language models, image generators, or scraped datasets)?
    \item[] Answer: \answerYes{}
    \item[] Justification: The release is limited to benchmark resources, processed episodes, schemas, logs, and evaluation code rather than a deployable clinical agent or a new high-risk model. The documentation states that the artifacts are for research evaluation only, includes source-use notes, and discourages clinical deployment claims.
    \Needspace{3\baselineskip}
    \item[] Guidelines:
    \begin{itemize}
        \item The answer \answerNA{} means that the paper poses no such risks.
        \item Released models that have a high risk for misuse or dual-use should be released with necessary safeguards to allow for controlled use of the model, for example by requiring that users adhere to usage guidelines or restrictions to access the model or implementing safety filters. 
        \item Datasets that have been scraped from the Internet could pose safety risks. The authors should describe how they avoided releasing unsafe images.
        \item We recognize that providing effective safeguards is challenging, and many papers do not require this, but we encourage authors to take this into account and make a best faith effort.
    \end{itemize}

\Needspace{4\baselineskip}
\item {\bf Licenses for existing assets}
    \item[] Question: Are the creators or original owners of assets (e.g., code, data, models), used in the paper, properly credited and are the license and terms of use explicitly mentioned and properly respected?
    \item[] Answer: \answerYes{}
    \item[] Justification: Appendix~\ref{app:asset_licenses} identifies the MedQA MIT license, the project CC BY-NC-SA 4.0 license, and the API service terms. The packages preserve the upstream notices and provide source and dependency attributions in the asset documentation.
    \Needspace{3\baselineskip}
    \item[] Guidelines:
    \begin{itemize}
        \item The answer \answerNA{} means that the paper does not use existing assets.
        \item The authors should cite the original paper that produced the code package or dataset.
        \item The authors should state which version of the asset is used and, if possible, include a URL.
        \item The name of the license (e.g., CC-BY 4.0) should be included for each asset.
        \item For scraped data from a particular source (e.g., website), the copyright and terms of service of that source should be provided.
        \item If assets are released, the license, copyright information, and terms of use in the package should be provided. For popular datasets, \url{paperswithcode.com/datasets} has curated licenses for some datasets. Their licensing guide can help determine the license of a dataset.
        \item For existing datasets that are re-packaged, both the original license and the license of the derived asset (if it has changed) should be provided.
        \item If this information is not available online, the authors are encouraged to reach out to the asset's creators.
    \end{itemize}

\Needspace{4\baselineskip}
\item {\bf New assets}
    \item[] Question: Are new assets introduced in the paper well documented and is the documentation provided alongside the assets?
    \item[] Answer: \answerYes{}
    \item[] Justification: The accompanying package documents the case records and their provenance, process derivatives, executable protocol, role-specific schemas, scoring and analysis code, configurations, and reproducibility instructions. The public code and dataset are linked in Section~\ref{sec:framework}.
    \Needspace{3\baselineskip}
    \item[] Guidelines:
    \begin{itemize}
        \item The answer \answerNA{} means that the paper does not release new assets.
        \item Researchers should communicate the details of the dataset\slash code\slash model as part of their submissions via structured templates. This includes details about training, license, limitations, etc. 
        \item The paper should discuss whether and how consent was obtained from people whose asset is used.
        \item At submission time, remember to anonymize your assets (if applicable). You can either create an anonymized URL or include an anonymized zip file.
    \end{itemize}

\Needspace{4\baselineskip}
\item {\bf Crowdsourcing and research with human subjects}
    \item[] Question: For crowdsourcing experiments and research with human subjects, does the paper include the full text of instructions given to participants and screenshots, if applicable, as well as details about compensation (if any)? 
    \item[] Answer: \answerYes{}
    \item[] Justification: Each of the two physicians received CNY 2,000 to review benchmark outputs and case conversions. The accompanying code package includes the full participant-facing instructions and both blank audit worksheets in \texttt{docs/physician\_review/}. No new patient data were collected.
    \Needspace{3\baselineskip}
    \item[] Guidelines:
    \begin{itemize}
        \item The answer \answerNA{} means that the paper does not involve crowdsourcing nor research with human subjects.
        \item Including this information in the supplemental material is fine, but if the main contribution of the paper involves human subjects, then as much detail as possible should be included in the main paper. 
        \item According to the NeurIPS Code of Ethics, workers involved in data collection, curation, or other labor should be paid at least the minimum wage in the country of the data collector. 
    \end{itemize}

\Needspace{4\baselineskip}
\item {\bf Institutional review board (IRB) approvals or equivalent for research with human subjects}
    \item[] Question: Does the paper describe potential risks incurred by study participants, whether such risks were disclosed to the subjects, and whether Institutional Review Board (IRB) approvals (or an equivalent approval/review based on the requirements of your country or institution) were obtained?
    \item[] Answer: \answerNA{}
    \item[] Justification: The institution determined that institutional ethics approval was not applicable to the paid expert audits of benchmark records and model outputs. No new patient data were collected.
    \Needspace{3\baselineskip}
    \item[] Guidelines:
    \begin{itemize}
        \item The answer \answerNA{} means that the paper does not involve crowdsourcing nor research with human subjects.
        \item Depending on the country in which research is conducted, IRB approval (or equivalent) may be required for any human subjects research. If you obtained IRB approval, you should clearly state this in the paper. 
        \item We recognize that the procedures for this may vary significantly between institutions and locations, and we expect authors to adhere to the NeurIPS Code of Ethics and the guidelines for their institution. 
        \item For initial submissions, do not include any information that would break anonymity (if applicable), such as the institution conducting the review.
    \end{itemize}

\Needspace{4\baselineskip}
\item {\bf Declaration of LLM usage}
    \item[] Question: Does the paper describe the usage of LLMs if it is an important, original, or non-standard component of the core methods in this research? Note that if the LLM is used only for writing, editing, or formatting purposes and does \emph{not} impact the core methodology, scientific rigor, or originality of the research, declaration is not required.
    \item[] Answer: \answerYes{}
    \item[] Justification: LLMs are core components of the evaluated doctor agents, optional MDT consultants, patient simulators, manager judges, case-structuring assistance, and rubric-based judging pipeline. These methodological uses are described separately from any LLM use only for writing, editing, or language polishing; all scientific claims and results are checked by the authors.
    \Needspace{3\baselineskip}
    \item[] Guidelines:
    \begin{itemize}
        \item The answer \answerNA{} means that the core method development in this research does not involve LLMs as any important, original, or non-standard components.
        \item Please refer to our LLM policy in the NeurIPS handbook for what should or should not be described.
    \end{itemize}

\end{enumerate}

\end{document}